\documentclass{bmvc2k}

\usepackage{import}
\usepackage{adjustbox}
\usepackage{booktabs}
\usepackage{multirow}
\usepackage{makecell}
\usepackage{graphicx}
\usepackage{amsmath}
\usepackage{amssymb}
\usepackage{url}
\usepackage{caption}
\usepackage{subcaption}
\usepackage{wrapfig}
\usepackage{bm}
\usepackage{algorithm}
\usepackage{algorithmic}
\usepackage{xcolor}

\title{Unsupervised Hashing with Similarity Distribution Calibration}

\addauthor{Kam Woh Ng}{kamwoh.ng@surrey.ac.uk}{1,2}
\addauthor{Xiatian Zhu}{xiatian.zhu@surrey.ac.uk}{1,3}
\addauthor{Jiun Tian Hoe}{jiuntian001@e.ntu.edu.sg}{4}
\addauthor{Chee Seng Chan}{cs.chan@um.edu.my}{5}
\addauthor{Tianyu Zhang}{macwish@hotmail.com}{6}
\addauthor{Yi-Zhe Song}{y.song@surrey.ac.uk}{1,2}
\addauthor{Tao Xiang}{t.xiang@surrey.ac.uk}{1,2}

\addinstitution{
  CVSSP, University of Surrey
}
\addinstitution{
  iFlyTek-Surrey Joint Research Centre on Artificial Intelligence, \\
  University of Surrey
}
\addinstitution{
  Surrey Institute for People-Centred Artificial Intelligence, \\ 
  University of Surrey
}
\addinstitution{
  Nanyang Technological University
}
\addinstitution{
  CISiP, Universiti Malaya
}
\addinstitution{
  GAC R\&D Center
}

\runninghead{Ng et al.}{Unsupervised Hashing with SDC}

\def\eg{\emph{e.g}\bmvaOneDot}

\def\ie{\emph{i.e}\bmvaOneDot}
\def\modelname{Similarity Distribution Calibration}
\def\shortname{SDC}

\newcommand{\Tstrut}{\rule{0pt}{1.5ex}}
\newcommand{\changed}[1]{#1}

\begin{document}

\maketitle

\begin{abstract}

Unsupervised hashing methods typically aim to preserve the similarity between data points in a feature space by mapping them to binary hash codes. However, these methods often overlook the fact that the similarity between data points in the continuous feature space may not be preserved in the discrete hash code space, due to the limited similarity range of hash codes.
The similarity range is bounded by the code length and can lead to a problem known as \textit{similarity collapse}. That is, the positive and negative pairs of data points become less distinguishable from each other in the hash space.
To alleviate this problem, in this paper a novel {\bf \modelname} (\shortname) method is introduced.
\shortname{} aligns the hash code similarity distribution 
towards a calibration distribution (\eg, beta distribution) with
sufficient spread across the entire similarity range, thus alleviating the similarity collapse problem. 
Extensive experiments show that our \shortname{} outperforms significantly the state-of-the-art alternatives on coarse category-level and instance-level image retrieval.
Code is available at \url{https://github.com/kamwoh/sdc}.
\end{abstract}

\section{Introduction} \label{sec:intro}

Hashing has been used extensively in real-world large-scale image retrieval systems. By converting continuous feature vectors into binary/discrete hash codes for indexing, hashing significantly reduces both computational cost and memory footprint.  Recent deep supervised learning to hash \cite{dpsh2016li, hashnet2017cao, dtq2018liu, greedyhash2018su, hcoh2018lin, csq2020yuan, dpn2020fan, dagh2019yang} have greatly outperformed conventional methods \cite{lsh1998indyk, spectral09weiss, bre2009kulis, itq2012gong, isotropic2012kong}. However, supervised hashing is limited in scalability due to its reliance on a large quantity of labeled training data. A natural solution is to use  {\em unsupervised hashing methods} instead, which do not require costly training data annotation. 

\begin{figure}[t!]
    \centering
    \begin{subfigure}[b]{0.32\linewidth}
        \centering
        \includegraphics[keepaspectratio=True, scale=0.34]{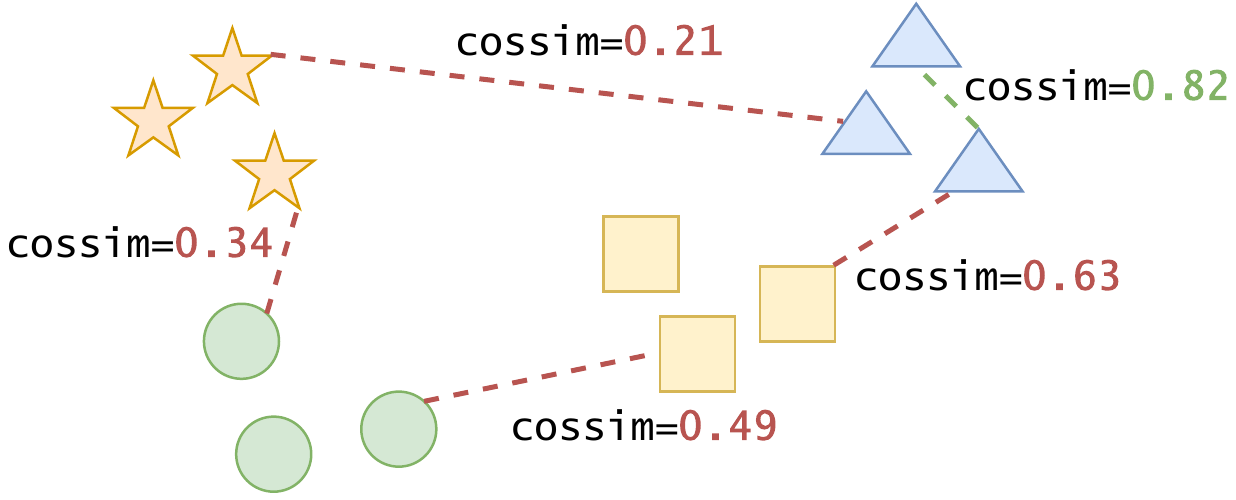}
        \caption{}
    \end{subfigure}
    \begin{subfigure}[b]{0.32\linewidth}
        \centering
        \includegraphics[keepaspectratio=True, scale=0.34]{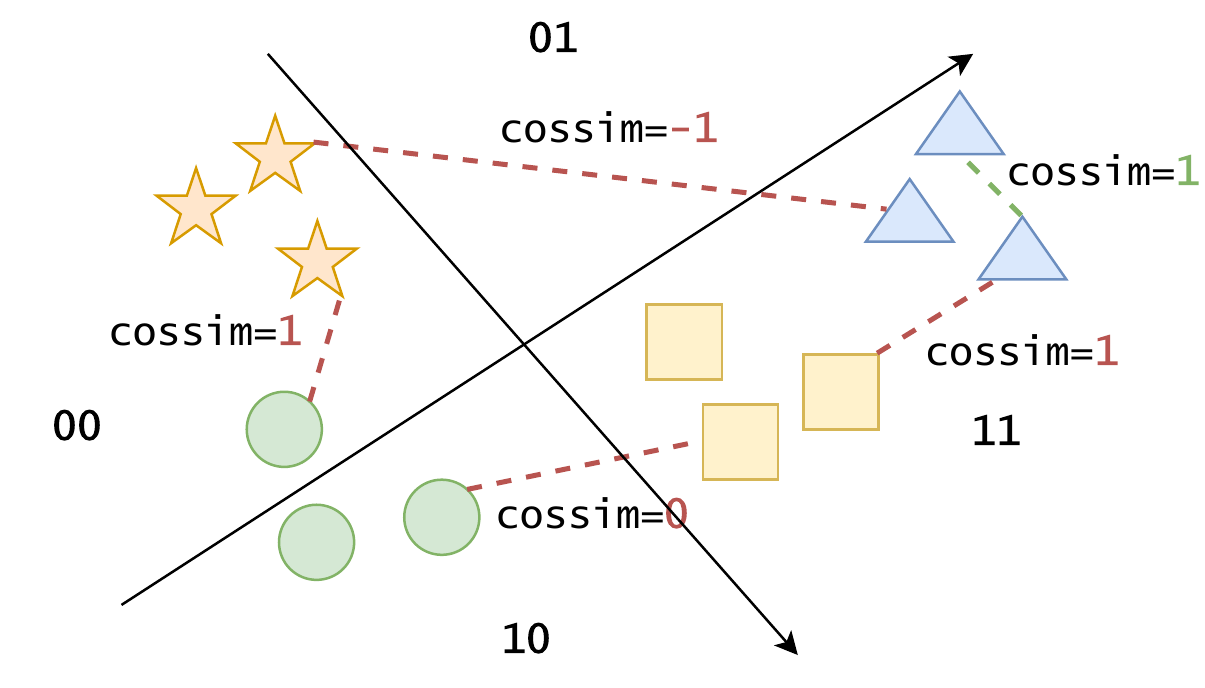}
        \caption{}
    \end{subfigure}
    \begin{subfigure}[b]{0.32\linewidth}
        \centering  
        \includegraphics[keepaspectratio=True, scale=0.34]{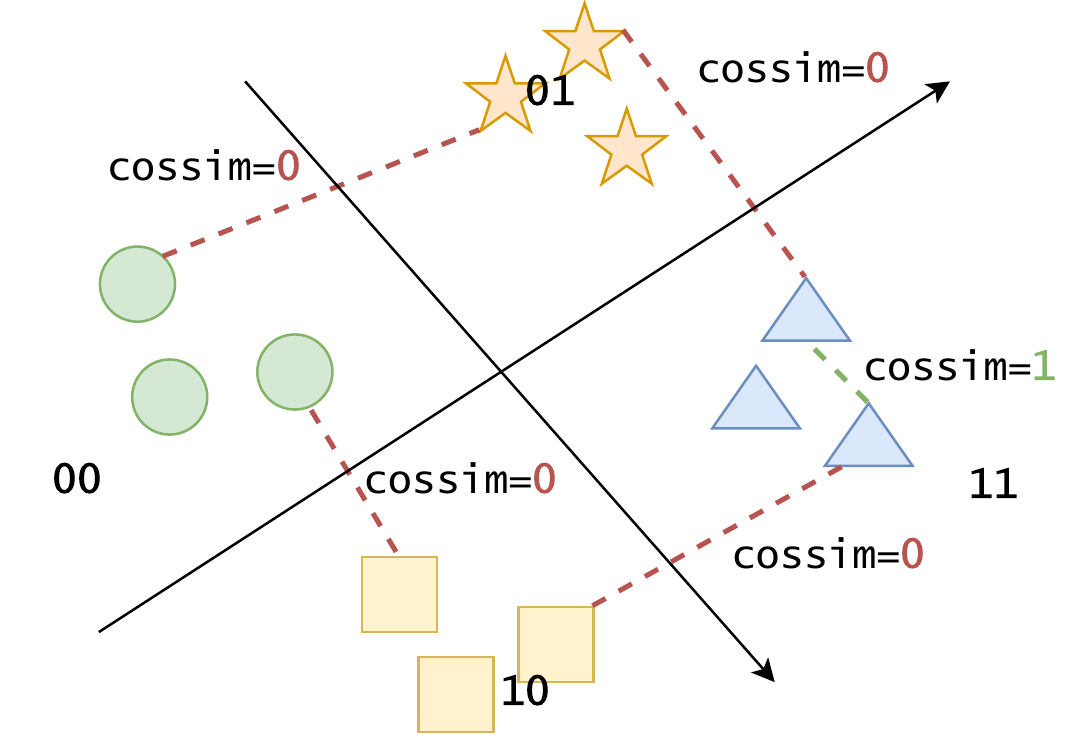}
        \caption{}
    \end{subfigure}
    \vspace{-0.3cm}
    \caption{
    \changed{\textbf{(a)} In the original feature space (before hashing), the cosine similarity values are continuous within the range of [-1, 1].
    \textbf{(b)} 
    After mapping to the hash code space with the conventional feature similarity preservation strategy,
    the similarity collapse phenomenon emerges due to the limited similarity range (only $K+1$ choices with $K$ the hash code length) in the Hamming space and the original similarity bias (\eg, negative pairs taking moderately positive similarity). 
    \textbf{(c)} This problem can be well alleviated 
    by our {\em \modelname{}} method even with limited cosine similarity values. See Fig.~\ref{fig:2bits} for the result on actual data.}
    } 
    \label{fig:bigteaser}
    \vspace{-0.4cm}
\end{figure}

The current state-of-the-art unsupervised hashing methods \cite{greedyhash2018su, bihalf2021li} are either based on {\em preserving individual pairwise similarities} between continuous feature vectors in the learned Hamming space \cite{bre2009kulis, are2018hu}. Compared to the alternative strategies (\eg, reconstruction \cite{binae2015car}, clustering \cite{pseudo2020yang}, pseudo-labels \cite{ssdh2017yang}, and contrastive learning \cite{cib2021qiu, wch2022yu, yu2022nsh}), pairwise similarity preservation is both easier to implement and more efficient, hence advantageous for large-scale  applications \cite{faiss2018joeff}.

However, we reveal that these similarity preservation-based hashing methods suffer from a \textit{\bf \em similarity collapse} problem (see Fig.~\ref{fig:bigteaser}b). That is, the hash code similarities of positive and negative pairs become inseparable.
There are two causes:
(i) The similarity distribution in the original continuous feature space is biased. In particular, most negative pairs take moderately positive similarity scores (Fig.~\ref{fig:bigteaser}a). 
(ii) The inherent difference in the ability of continuous feature space and discrete hash code space to accurately preserve similarity between data points.
Specifically, the similarities between any two hash codes are of a fixed set of values determined by the code length (\ie, limited capacity), whilst the original feature similarities are continuous (\ie, unlimited capacity).
With the limited range of similarity scores in the hash code space, the hashing process is given little chance to recover from the collapsing positive and negative feature similarity scores in the Hamming space (Fig.~\ref{fig:bigteaser}b), resulting in inferior retrieval results.  

To alleviate this similarity collapse problem, in this work a novel {\em \modelname{}} ({\bf \shortname}) method is introduced. Instead of preserving the original pairwise similarity scores individually, we \changed{regularize} the hash code similarity distribution 
as a whole against a pre-defined calibration distribution (\eg, beta distribution) with sufficient capacity range.
Due to this stretching effect, the learned Hamming space is no longer restricted severely by the original biased similarity distribution as in the existing methods.
This enables the limited similarity range of Hamming space to be better leveraged, resulting in improved performance (Fig.~\ref{fig:bigteaser}c).

We make the following {\bf contributions}: 
(i) We reveal the fundamental similarity collapse problem suffered by existing pairwise similarity preservation-based unsupervised hashing methods.
(ii) To address this problem, we propose a \modelname{} (\shortname{}) method by alleviating the severe restriction imposed by the original biased similarity scores. 
(iii) Extensive experiments validate the superiority of our \shortname{} over state-of-the-art alternatives on four category-level and three instance-level image retrieval benchmarks.

\section{Related Work} \label{sec:related_work}

Although earlier hashing methods \cite{lsh1998indyk, spectral09weiss, itq2012gong, pca1986jolliffe, angular2012gong, hammingmetric2012mohammad, isotropic2012kong, ksh2012liu} are easy to apply in practice, their performance is typically inferior to more recent deep learning counterparts. Deep supervised hashing methods \cite{dpsh2016li, hashnet2017cao,  hcoh2018lin, hmoh2020lin, greedyhash2018su, dpn2020fan, dtsh2016wang, csq2020yuan, dagh2019yang} usually achieve better performance over unsupervised ones by using additionally the semantic class labels. However, they are limited in scalability as class label annotation is costly and even impossible in extreme cases (\eg, rare objects).
Without this constraint, unsupervised methods are thus more scalable.
Existing unsupervised hashing methods can be categorized into the following groups: similarity preservation \cite{bre2009kulis, lph2014zhao, dgh2014liu, deepbit2016lin, uth2017huang, are2018hu, greedyhash2018su, sadh2018shen, usdh2019sheng, nph2019li, bihalf2021li}, generative model  \cite{sgh2017dai, bgan2018song, bingan2018zieba, hashgan2018cao}, reconstruction \cite{tbh2020shen, binae2015car, dvb2019shen, chen2018learning}, pseudo-labeling \cite{ssudh2018yang, distill2019yang, udhpa2019zhu, udth2019gu, du3h2020zhang, cimon2021luo}, clustering \cite{udhpa2019zhu, pseudo2020yang, du3h2020zhang, dcuh2021yu} and contrastive learning \cite{pseudo2020yang, cib2021qiu, lthsort2022shen}.

Among these, similarity preservation-based unsupervised hashing methods achieve the current state-of-the-art performance in image retrieval.
They are also simple in design and efficient computationally. \changed{The general idea is to learn a set of hash codes based on the feature similarity information. This can be achieved either with projection \cite{pca1986jolliffe, itq2012gong}, inferring pseudo-labels \cite{ssudh2018yang, distill2019yang, cimon2021luo} or direct preserving the feature  similarity \cite{bre2009kulis, are2018hu, greedyhash2018su, bihalf2021li}.
Recent contrastive learning based hashing \cite{cimon2021luo, cib2021qiu, wch2022yu, yu2022nsh} further push the performance.
Beyond all these learning approaches,
we focus on the similarity collapse problem
overlooked by these prior studies.
With ground-truth labels, 
similarity collapse can be suppressed
by explicitly constraining positive and negative pairs to have small and large Hamming distances
\cite{mihash2017fatih, mihashjournal2019cakir, kemertas2020rankmi}. %
This, however, does not fit unsupervised hashing.
By imposing similarity distribution prior, our SDC elegantly eliminates the need for training labels.
%
 }

\section{Methodology} \label{sec:methodology}

To obtain a hash code $b \in \{-1,+1\}^K$ with $K$ bits, we need a hash function $h$ as:
\begin{align} \label{eq:hashfunction}
    \mathbf{b} = h(\mathbf{x})=\texttt{sign}(\phi(\mathbf{x})),
\end{align}
where $\phi:\mathbf{x} \rightarrow \mathbf{f} \in \mathbb{R}^K$ is a (non-)linear mapping function compressing a $d$-dimensional feature vector $\mathbf{x} \in \mathbb{R}^d$ into a $K$-dimensional continuous code $\mathbf{f}$. $h$ is learned by optimizing an objective function $\mathcal{L}$. At test time for image retrieval, the Hamming distances between a query code, $\mathbf{b}_p$, and the gallery codes, $\mathbf{b}_q$, of a database can be computed as:
\begin{align} \label{eq:hamming_theta}
    \mathcal{D}_h(\mathbf{b}_p, \mathbf{b}_q) = \frac{K}{2} (1 - \cos{\theta_{pq}}),
\end{align}
where $\cos{\theta_{pq}}=\cos(\mathbf{b}_p,\mathbf{b}_q)$ is the cosine similarity between $\mathbf{b}_p$ and $\mathbf{b}_q$.

It is not differentiable with Eq. \eqref{eq:hashfunction} in a hashing objective $\mathcal{L}$ due to the non-differentiable $\texttt{sign}$ function. A straightforward solution is to remove the $\texttt{sign}$ function, whilst minimizing the quantization error between $\mathbf{f}$ and its hash code $\mathbf{b}$ during training \cite{lthsurvey2020luo} as: 
\begin{align} \label{eq:cosquan}
    \mathcal{L}_{q} = \frac{1}{N} \sum^{N}_{i=1} (1 - \cos{\theta_{i}}), \;\; \text{and} \;\; \cos{\theta_{i}}=\cos(\mathbf{f}_i,\mathbf{b}_i),
\end{align}
where $\cos{\theta_{i}}$ is the cosine similarity between the continuous code $\mathbf{f}_i$ and the hash code counterpart $\mathbf{b}_i=\texttt{sign}(\mathbf{f}_i)$ of $i$-th sample, and $N$ specifies the training set size.
This enables differentiable end-to-end hashing without a straight-through estimator \cite{ste2013bengio, greedyhash2018su} or continuous relaxation \cite{hashnet2017cao}.
Note, although the continuous codes $\mathbf{f}$ are involved in learning, we describe the learning process directly with hash codes $\mathbf{b}$ hereafter for convenience.



\subsection{Hashing by Conventional Similarity Preservation} \label{sec:prob}

Prior arts \cite{lsh1998indyk, pca1986jolliffe, bre2009kulis, itq2012gong, are2018hu, greedyhash2018su, bihalf2021li} preserve the pairwise similarities
of the original continuous feature space during hashing. 
The loss function is often formulated as:
\begin{align} \label{eq:preserveloss}
    \mathcal{L}_{\text{p}}=\frac{1}{|\mathcal{N}|}\sum_{(i,j)\in\mathcal{N}}|t_{(i,j)}-s_{(i,j)}|^p,
\end{align}
where $p=2$, $t_{(i,j)}=\cos(\mathbf{x}_i,\mathbf{x}_j)$ is the similarity reconstruction target for $\mathbf{x}_i$ and $\mathbf{x}_j$ drawn from a training set $\mathbf{X} \in \mathbb{R}^{N \times d}$, $\mathcal{N}$ is a set of selected sample pairs, $|\mathcal{N}|$ is the set cardinality, and $s_{(i,j)}=\cos(\mathbf{b}_i, \mathbf{b}_j)$ is the similarity of hash codes. Each hash code is obtained with a hash function $\mathbf{b} = h(\mathbf{x})$ (Eq. \eqref{eq:hashfunction}).

\begin{wrapfigure}{r}{0.48\textwidth}
    \centering
    \vspace{-0.7cm}
    \includegraphics[scale=0.22, keepaspectratio=True]{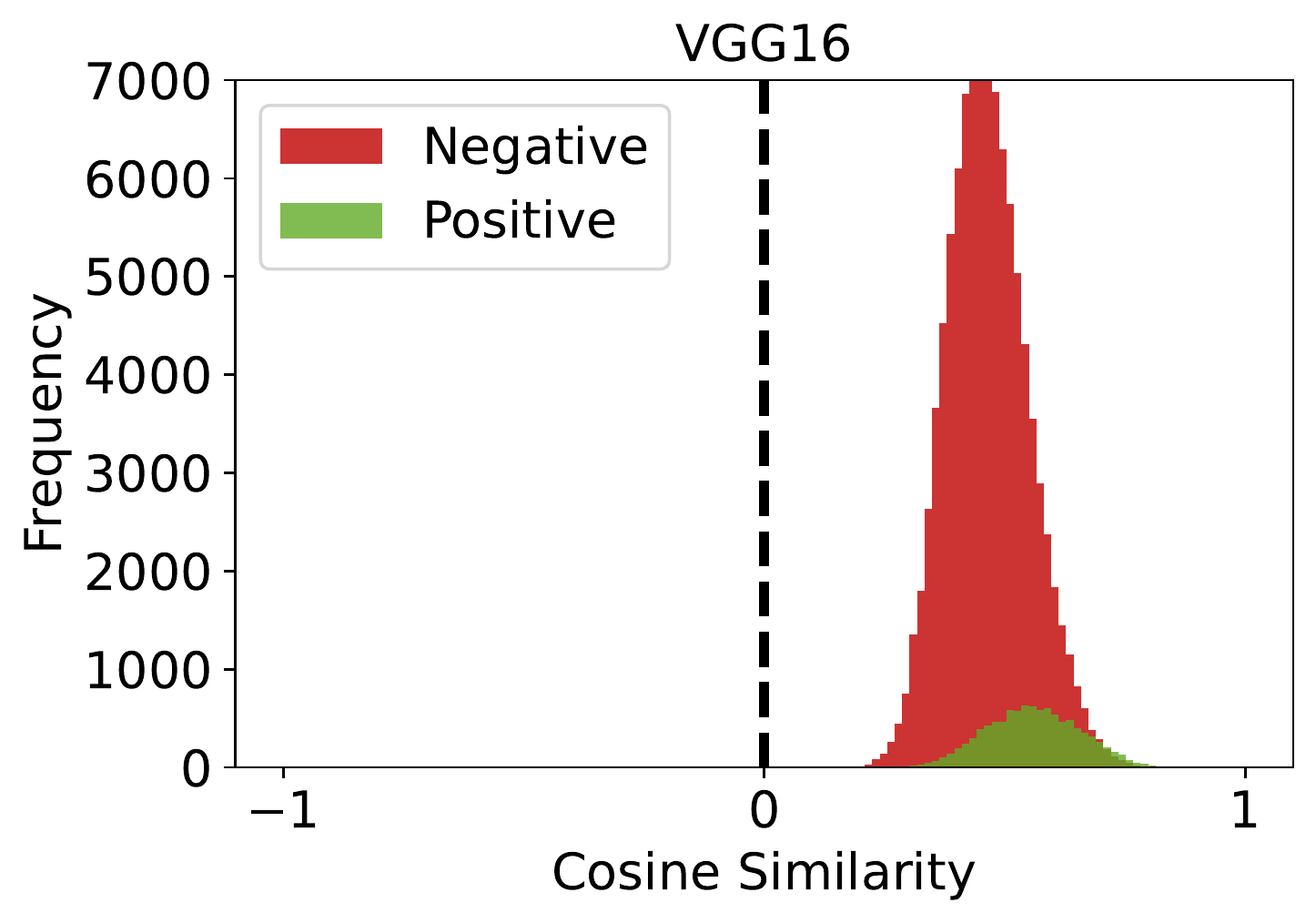}
    \includegraphics[scale=0.22, keepaspectratio=True]{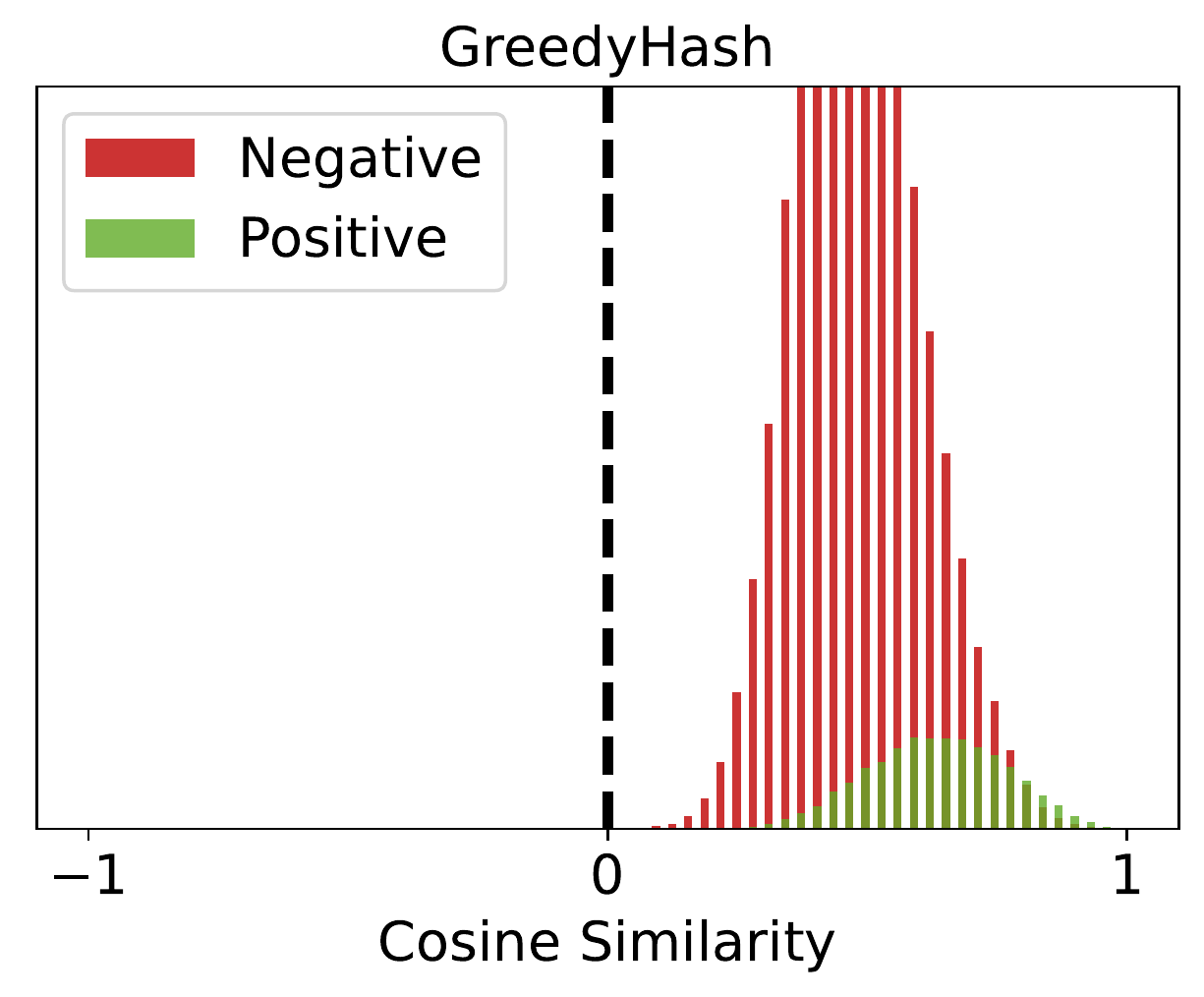}
    \caption{Pairwise similarity distribution of CIFAR10 using 100k randomly chosen negative pairs and 10k positive pairs. \textbf{(Left)} VGG16 features. \textbf{(Right)} 64-bits hash codes of GreedyHash \cite{greedyhash2018su}. For easy explanation, positive and negative labels are included (\ie descriptive purpose only), and were not used during the actual unsupervised training.}
    \label{fig:disthist_orig_and_ugh}
    \vspace{-0.4cm}
\end{wrapfigure}

As discussed earlier, similarity preservation based unsupervised hashing methods suffer from a {\bf \em similarity collapse} problem, as indicated by the severe overlapping in the hash code similarity scores of positive and negative pairs (Fig. \ref{fig:bigteaser}b). 
Intuitively, this would lead to suboptimal retrieval performance.

As a concrete example, we examine the pairwise similarity distribution of CIFAR10 in the Hamming space. From Fig.~\ref{fig:disthist_orig_and_ugh} we observe that the distribution of hash code similarities is mainly concentrated in the positive region. This is because similarity preservation (\ie, Eq.~\eqref{eq:preserveloss}) would directly inherit the similarity bias of the original feature space (VGG-16 features in this case).
As a result, the similarity range of Hamming space is
leveraged only at a limited degree, giving rise to the similarity collapse problem. 

\subsection{Similarity Distribution Calibration} \label{sec:maxneigh}


\begin{figure}[t!]
    \centering
    \includegraphics[keepaspectratio=True, width=\linewidth]{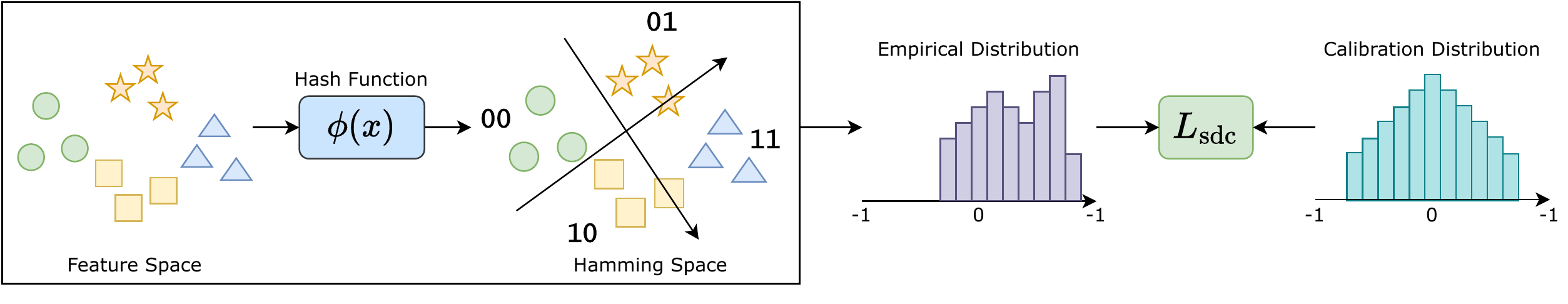}
    \vspace{-0.6cm}
    \caption{
    \changed{Pipeline of our proposed \modelname{} (\shortname). 
    We first map the original features into the Hamming space through a learnable hash function. Next, we construct the empirical hash code similarity distribution.
    Finally, we minimize the Wasserstein distance ($L_\text{sdc}$) between the empirical distribution and a calibration distribution.}
    } 
    \label{fig:illustration}
    \vspace{-0.2cm}
\end{figure}

{\em \modelname{}} (\shortname{}) is designed particularly for alleviating the similarity collapse problem. 
The idea is to align the {empirical} hash code similarity distribution of the training data with a calibration distribution with sufficient spread across the entire similarity range. 
To measure the discrepancy between two probability distributions for similarity calibration, we adopt the Wasserstein distance with  an elegant solution based on {\em inverse Cumulative Distribution Function} (iCDF) \cite{rabin2011wasserstein, ramdas2017wasserstein}. 
Formally, we consider the hash code similarity ${s}$ as a random variable with the iCDF $F$ \changed{conditioned on the {\em feature similarities} $t$}. Our Wasserstein distance-based calibration is formulated as:
%
\begin{align} \label{eq:emd}
    \int^{1}_{0} |{F}(z)-C(z)|dz,
\end{align}
where $z$ is the quantile with the interval of $[0, 1]$,
and $C$ is the iCDF of the calibration distribution.
The pipeline of SDC is depicted in Fig.~\ref{fig:illustration}.


\vspace{0.1cm}
\noindent \textbf{Approximation.} 
We can estimate $F$ by collecting the pairwise similarities of hash codes and sorting them in the ascending order of $t$.
Concretely, we evenly divide the probability range $[0,1]$
into $|\mathcal{N}|$ bins and then aggregate per-bin calibration as:
\begin{align} \label{eq:emdapprox}
    \mathcal{L}_\text{sdc} = \frac{1}{|\mathcal{N}|} \sum_{s_i \in \mathcal{N}} ~ \bigr|s_i - C(\frac{2i-1}{2|\mathcal{N}|})\bigr|,
\end{align}
where $s_{i}$ is $i$-th sorted hash code pairwise similarity. 
%
\changed{$\mathcal{L}_\text{sdc}$ has a similar form to $\mathcal{L}_\text{p}$. 
$\mathcal{L}_\text{p}$ with $p=1$ is essentially minimizing the Wasserstein distance between original feature similarity and hash code similarity distribution. 
Sorting by the order of $t$ conditions the iCDF $F$,
balancing between feature similarity preservation and Hamming similarity range usage.
}

\begin{figure}[t]
    \centering
    \begin{subfigure}[b]{0.245\textwidth}
        \centering
        \includegraphics[scale=0.23, keepaspectratio=True]{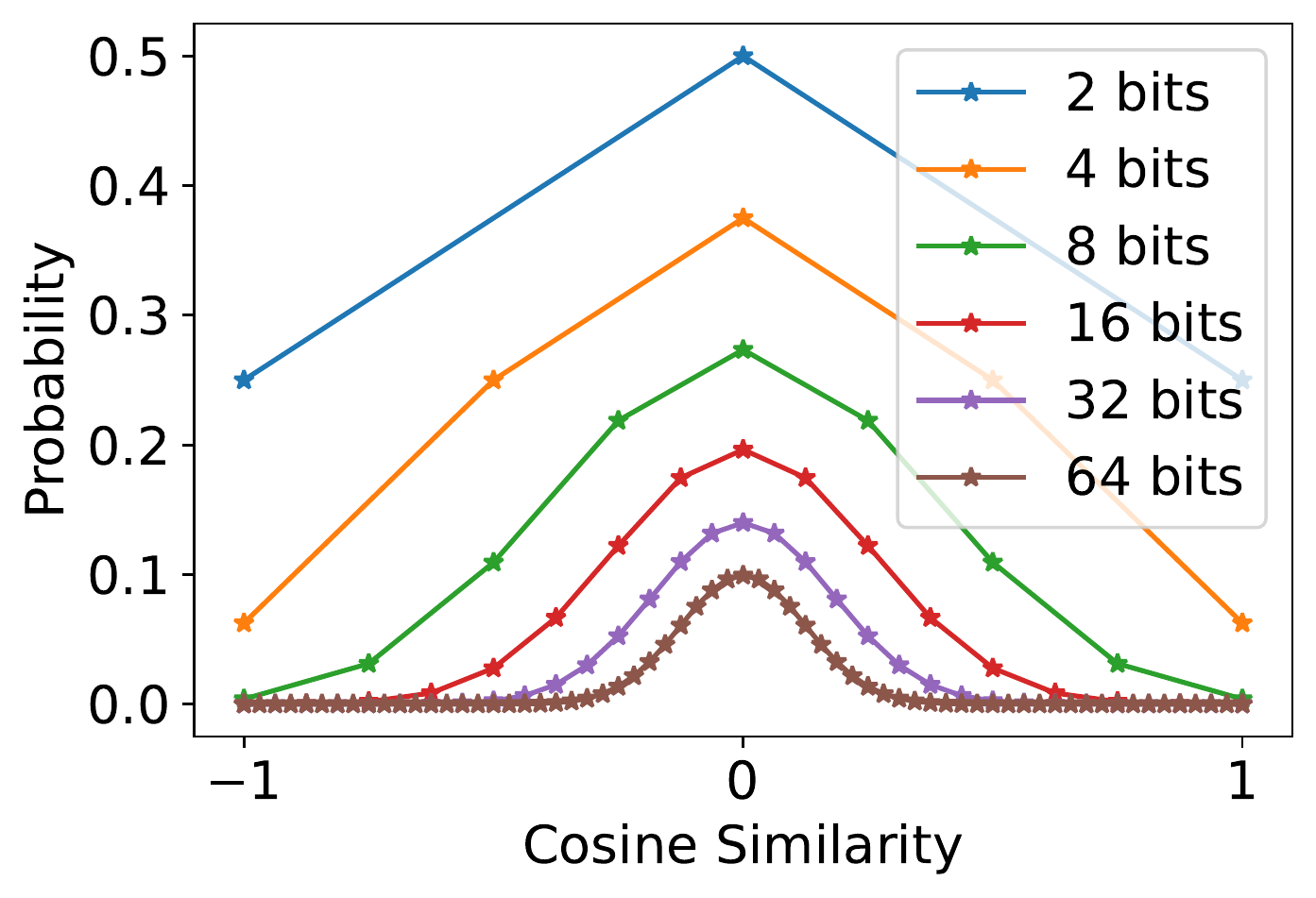}
        \caption{}
        \label{fig:bin_pmf}
    \end{subfigure}
    \begin{subfigure}[b]{0.245\textwidth}
        \centering
        \includegraphics[scale=0.23, keepaspectratio=True]{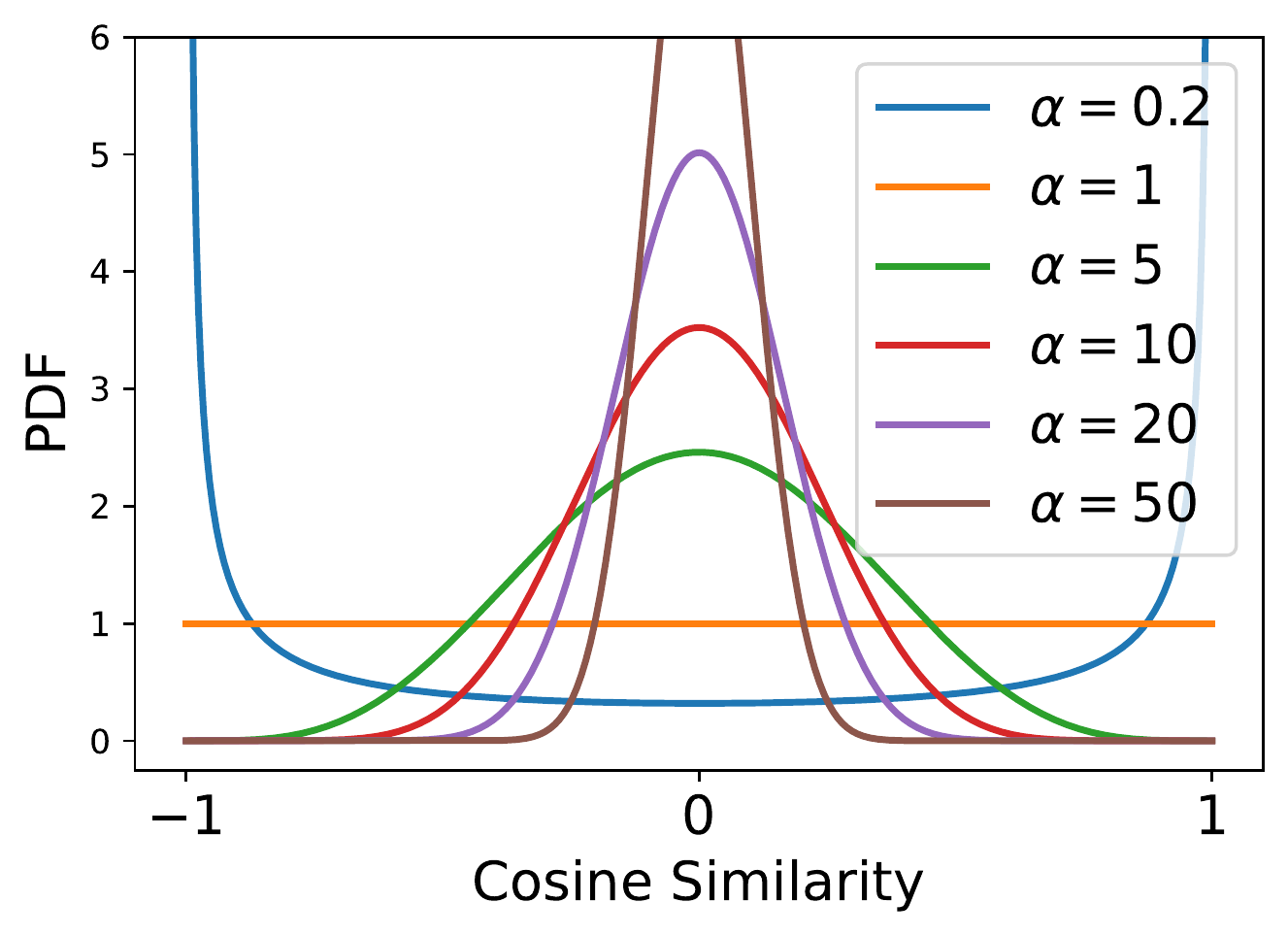}
        \caption{}
        \label{fig:beta_pdf}
    \end{subfigure}
    \begin{subfigure}[b]{0.245\textwidth}
        \centering
        \includegraphics[scale=0.23, keepaspectratio=True]{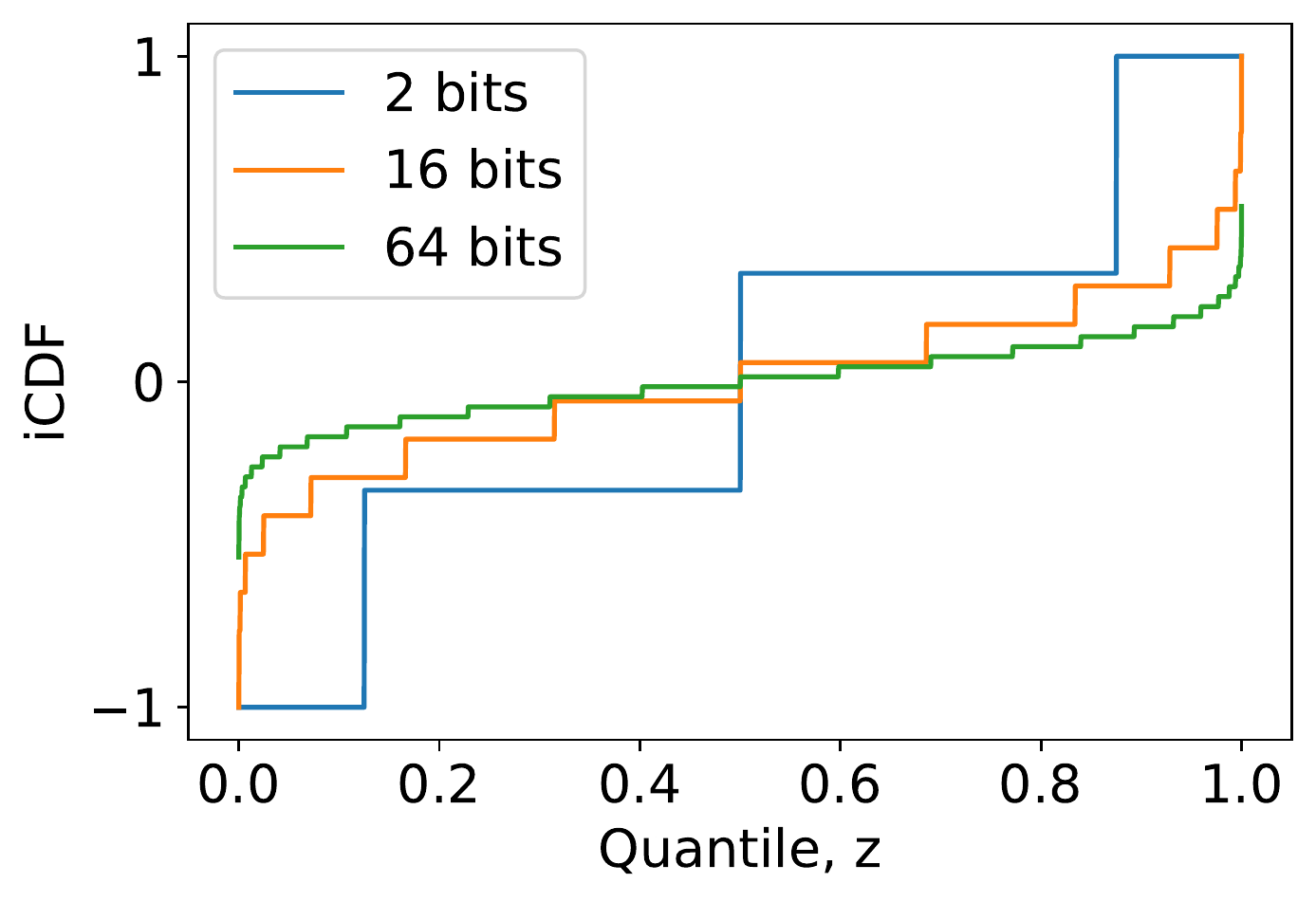}
        \caption{}
        \label{fig:bin_icdf}
    \end{subfigure}
    \begin{subfigure}[b]{0.245\textwidth}
        \centering
        \includegraphics[scale=0.23, keepaspectratio=True]{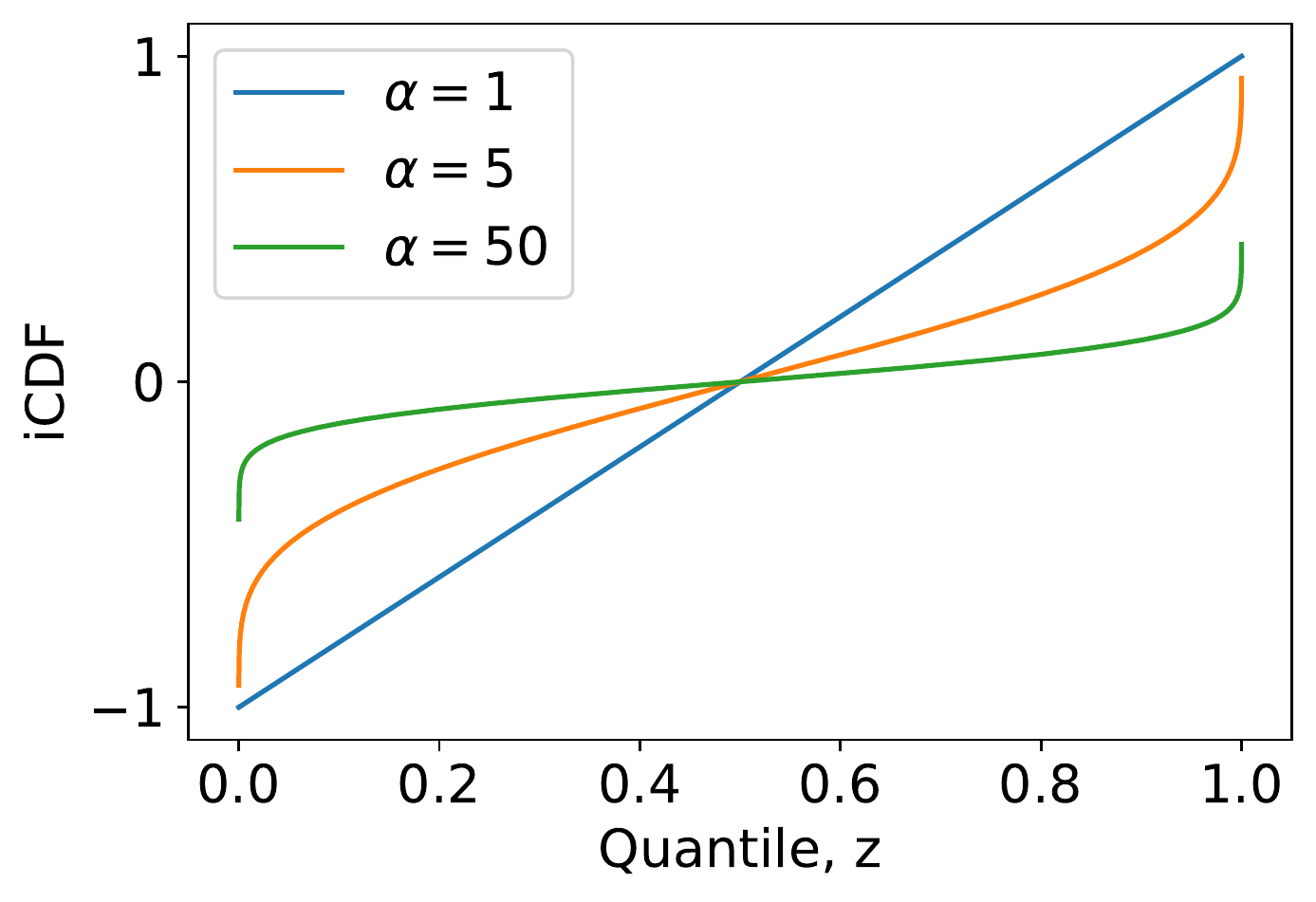}
        \caption{}
        \label{fig:beta_icdf}
    \end{subfigure}
    \vspace{-0.6cm}
    \caption{
    (\textbf{a)} Probability mass function (PMF) of $\texttt{B}(K,0.5)$ and \textbf{(b)} probability density function (PDF) of $\texttt{Beta}(\alpha,\beta)$ distribution with different $\alpha/\beta$ values. \textbf{(c)} Inverse cumulative density function (iCDF) of $\texttt{B}(K,0.5)$ and \textbf{(d)} $\texttt{Beta}(\alpha,\beta)$ distribution with different $\alpha/\beta$ values. 
    Note that we set $\alpha =\beta$ for symmetric PDF. 
    }
    \label{fig:pmf_pdf_icdf}
    \vspace{-0.3cm}
\end{figure}

\vspace{0.1cm}
\noindent \textbf{Instantiation.} In general, any distribution with sufficient capacity spread is suited for calibration distribution.
As an instantiation, we consider {\bf\em beta distribution}, $\texttt{Beta}(\alpha, \beta)$ with $\alpha$ and $\beta$ the two positive shape parameters. We set $\alpha=\beta$ for simpler symmetric beta distribution. \changed{$\texttt{Beta}(\alpha, \beta)$ is chosen as its iCDF is bounded to $[0, 1]$, making it easy to be transformed to the target similarity range (\eg, $[-1, 1]$ for cosine similarity in our case) when minimizing $\mathcal{L}_\text{sdc}$.}%
%
%
There is no prior knowledge about the optimal parameter value.  
However, as illustrated in Fig.~\ref{fig:beta_pdf}, the shapes of probability density functions (PDF) over different parameter values all meet the requirements \changed{(\ie, fully utilizing the entire similarity range)} as calibration distribution.
Empirically, we find that $\alpha=\beta=5$ works generally \changed{(approximating a Gaussian distribution of $N(\mu=0, \sigma=0.3)$).}

\vspace{0.1cm}
\noindent \textbf{Hash buckets perspective.} We interpret the \shortname{}
from the hash buckets perspective.
It is assumed that an optimal hash function should encode similar items with the same hash code (preserved similarity) and fully utilize the Hamming space (decorrelated and balanced bit) \cite{spectral09weiss}. In an ideal inverted file system \cite{pqnns2010jegou}, $K$-bits hash codes can form $2^K$ hash buckets all of which are used at equal size. 
Thus any two hash codes can be sampled uniformly.

For the balanced bits case, the probability that one bit differs is $0.5$, and that $d$ bits differ 
is $(0.5)^d (0.5)^{K-d}$. 
There are ${K \choose d}$ variants for $d$ bits being different.
%
Thus, the probability that the Hamming distance between two uniformly sampled $K$-bits hash codes equals to $d$ is:
\begin{align} \label{eq:probdistance}
    {K \choose d} (0.5)^d (0.5)^{K-d} = \frac{1}{2^K}{K \choose d}.
\end{align}
This is equivalent to the probability mass function of a binomial distribution $\texttt{B}(K,0.5)$. We plot the result in Fig.~\ref{fig:bin_pmf} and the iCDF of Eq.~\eqref{eq:probdistance} in Fig.~\ref{fig:bin_icdf}. We see that as $K$ varies from $2$ to $64$, the similarity distribution is similar to the beta distribution with $\alpha=\beta \rightarrow \infty$ (see Fig.~\ref{fig:beta_icdf}).
This means that for learning an optimal hash function, we should produce hash codes with their pairwise similarity distribution similar to a binomial distribution. The case that Hamming space is not fully used suggests imbalanced hash bucket sizes,
meaning a biased similarity distribution and similarity collapse emerges.

 

\noindent \textbf{Remarks.} The feature similarity scores are used to sort the hash code counterparts while constructing the iCDF.
However, unlike the conventional strategy preserving {\em individual} pairwise similarity scores rigidly,
our \shortname{} leverages the distribution of feature similarity scores {\em holistically}. {We observe that pairwise feature similarities could vary over mini-batches.
During calibration, we apply sorting to rank them before aligning their corresponding hash code similarities with the prior distribution. As a result, our method does not conduct one-to-one alignment between feature similarity and prior distribution. This property could be understood as a type of stochastic noise during optimization,
in the spirit of SGD.
\changed{Critically, SDC maintains the cluster structure of original features better (Fig. \ref{fig:2bits}), making it more discriminative for image retrieval.}

\subsection{Overall Learning Objective} \label{sec:hashlearn}

For model training, we deploy the overall objective loss function as $\mathcal{L} = \mathcal{L}_{\text{sdc}} + \lambda_{q} \mathcal{L}_{q} + \lambda_{\text{cl}} \mathcal{L}_{\text{cl}}$
%
%
where $\mathcal{L}_{q}$ is the quantization loss (Eq. \eqref{eq:cosquan}), $\mathcal{L}_{\text{cl}}$ is contrastive learning loss \cite{simclr2020chen} as adopted by recent contrastive hashing methods \cite{cib2021qiu, yu2022nsh, wch2022yu}, both $\lambda_{\text{cl}}$ and $\lambda_{q}$ are hyper-parameters. 
We simply set $\lambda_{q} = 1$ and $\lambda_{\text{cl}} = 1$ unless mentioned otherwise. See supplementary material for the detailed algorithm.


\section{Experiments} \label{sec:exp}


\noindent \textbf{Datasets. } 
We consider both coarse category-level and fine-grained instance-level image retrieval tasks in our experiments.
Following \cite{greedyhash2018su, dpn2020fan, bihalf2021li, tbh2020shen, cib2021qiu}, we use 4 category-level datasets: i) \textbf{CIFAR-10} \cite{cifar102009krizhevsky}, ii) \textbf{NUS-WIDE} \cite{nuswide2009chua}, and iii) \textbf{MS-COCO} \cite{coco2014lin}. With an ImageNet pre-trained model,  we also choose iv) \textbf{ImageNet100} (a subset of ImageNet \cite{imagenet2009deng} as first used by \cite{hashnet2017cao} and later by supervised deep hashing works), which was ignored by previous unsupervised hashing works. 
For evaluating instance-level retrieval tasks,
three popular datasets are chosen including i) \textbf{GLDv2} \cite{gldv22020weyand}, ii) $\mathcal{R}$\textbf{Oxf} \cite{oxford2007philbin, roxfparis2018filip}, and iii) $\mathcal{R}$\textbf{Paris} \cite{paris2008philbin, roxfparis2018filip}. 

\noindent \textbf{Evaluation metrics.} Following previous works \cite{cib2021qiu, greedyhash2018su, bihalf2021li}, we measure the model performance with mean Average Precision (mAP) at top 1000 (mAP@1K) for single-labeled datasets (\ie, ImageNet100 and CIFAR10), while top 5000 (mAP@5K) for multi-labeled datasets (\ie, NUS-WIDE and MS-COCO). Note, for instance-level retrieval tasks, we follow the evaluation protocol of \cite{delg2020cao,ortho2021jk} and use mAP@100 as the evaluation metric. For statistical stability, we run 3 trials for each experiment and report the average of per-trial best results on the validation set. 

\vspace{0.1cm}
\noindent \textbf{Competitors.} 
We consider 3 classic unsupervised hashing methods \cite{lsh1998indyk, spectral09weiss, itq2012gong}, and 7 recent state-of-the-art unsupervised deep hashing methods \cite{ssudh2018yang, greedyhash2018su, tbh2020shen, bihalf2021li, cib2021qiu, yu2022nsh, wch2022yu}.

\noindent \textbf{Implementation details.} 
For fair comparisons, we follow the existing experimental protocol \cite{greedyhash2018su, bihalf2021li, tbh2020shen, cib2021qiu}.
\changed{We use the same pretrained feature extractor (\textit{frozen})
as the competitors (ResNet50 \cite{resnet2015he} for NSH \cite{yu2022nsh}, ViT-B/16 \cite{vit2020alexey} for WCH \cite{wch2022yu}, and VGG-16 \cite{vgg2015karen} for the rest).}
%
%
We test three common code lengths: 16, 32, and 64 bits.
We train all the compared methods using Adam \cite{adam2014kingma} optimizer for 100 epochs with a learning rate of $0.0001$ and a batch size of 64, with a single exception in TBH \cite{tbh2020shen}, for which a batch size of 400 is used and 1000 epochs are required.
Note that we re-implemented \changed{most} competing methods based on the original released codes.
Our reimplementation can reproduce the reported performances under the original setting. 
This allows us to evaluate all the models fairly
under a single setting sharing the same datasets, testing protocols, and network architectures.
More experimental details including the training/query/gallery splits for each dataset and implementation details \changed{including hyperparameters} are given in the supplementary material.

\subsection{Comparative Results} \label{sec:performance}

{\bf Coarse category-level retrieval results.} 
We report the retrieval results of our \shortname{} and prior art alternatives on three datasets in Table \ref{tab:retrieval_performance_unsupervised}.
\changed{We see that \shortname{} outperforms 
consistently all the competitors,
suggesting the importance of similarity collapse
as revealed in this work.
%
The precision-recall (PR) curves in Fig.~\ref{fig:pr_curve} show that our \shortname{} (blue curves) excels across different recall rates, especially at low bit cases (\ie, 16-bits).
}


\begin{table*}[t]
    \centering
    \begin{adjustbox}{max width=\textwidth}
        \begin{tabular}{|l|c|ccc|ccc|ccc|ccc|}
        \hline 
        \multirow{2}{*}{Methods} & \multirow{2}{*}{{Reference}} & \multicolumn{3}{c|}{\makecell[c]{\Tstrut CIFAR10}} & \multicolumn{3}{c|}{\makecell[c]{\Tstrut ImageNet100}} & \multicolumn{3}{c|}{\makecell[c]{\Tstrut NUSWIDE}} & \multicolumn{3}{c|}{\makecell[c]{\Tstrut MS-COCO}} \\ \cline{3-14} \Tstrut
         & & 16 & 32 & 64 & 16 & 32 & 64 & 16 & 32 & 64 & 16 & 32 & 64 \\ \hline \Tstrut
         \textit{VGG16} & & & & & & & & & & & & &  \\
        LsH \cite{lsh1998indyk} & STOC'98 & 23.9 & 29.6 & 37.6 & 14.7 & 29.7 & 48.7 & 51.0 & 59.3 & 67.1 & 45.2 & 51.6 & 59.8 \\
        SH \cite{spectral09weiss} & NeurIPS'08 & 41.8 & 42.1 & 43.5 & 35.1 & 50.9 & 60.9 & 63.0 & 60.9 & 64.0 & 59.4 & 64.8 & 66.2 \\ 
        ITQ \cite{itq2012gong} & TPAMI'12 & 46.8 & 51.3 & 54.4 & 45.5 & 62.1 & 72.7 & 73.2 & 75.0 & 77.1 & 67.6 & 72.9 & 75.4 \\
        SSDH \cite{ssudh2018yang} & IJCAI'18 & 41.0 & 39.6 & 38.5 & 32.3 & 40.1 & 44.6 & 66.8 & 67.8 & 66.7 & 53.9 & 56.7 & 57.4 \\ 
        GreedyHash \cite{greedyhash2018su} & NeurIPS'18 & 44.9 & 51.9 & 55.7 & 54.4 & 68.7 & 74.7 & 70.0 & 76.2 & 79.3 & 66.8 & 73.2 & 77.4 \\ 
        TBH \cite{tbh2020shen} & CVPR'20 & 48.2 & 50.2 & 50.7 & 42.9 & 44.5 & 48.3 & 75.8 & 77.8 & 78.5 & 68.8 & 72.6 & 74.8 \\ 
        CIBHash$^\dagger$ \cite{cib2021qiu} & IJCAI'21 & {56.2} & 59.2 & 61.2 & 63.9 & 71.4 & 74.6 & 77.1 & 79.7 & 80.9 & 73.3 & 77.0 & 78.5 \\ 
        BiHalf \cite{bihalf2021li} & AAAI'21 & 54.7 & 58.1 & 60.6 & 60.7 & 71.2 & 76.0 & 77.4 & 80.1 & 81.9 & 71.2 & 75.6 & 78.0 \\
        \Tstrut
        \bf \shortname$^\dagger$ & \bf Ours & \bf 59.8 & \bf 64.0 & \bf 66.3 & \bf 72.8 & \bf{78.5} & \bf{80.6} & \bf{80.7} & \bf{82.3} & \bf{83.4} & \bf 76.9 & \bf 79.8 & \bf 81.2 \\
        \hline \Tstrut
        \textit{ResNet50} & & & & & & & & & & & & &  \\
        NSH$^\dagger$ \cite{yu2022nsh} & IJCAI'22 & 70.6* & 73.3* & 75.6* & - & - & - & 75.8* & 81.1* & 82.4* & 74.6* & 77.4* & 78.3* \\ 
        \bf \shortname$^\dagger$ &  \bf Ours & \bf{74.2} & \bf{75.8} & \bf 78.4 & \bf 80.7 & \bf 83.8 & \bf 85.7 & \bf 81.2 & \bf 83.2 & \bf 84.2 & \bf 78.3 & \bf 81.1 & \bf 82.6  \\
        \hline \Tstrut
        \textit{ViT-B/16} & & & & & & & & & & & & &  \\
        WCH$^\dagger$ \cite{wch2022yu} & ACCV'22 & 77.5 & 79.3 & 80.6 & 69.4 & 76.9 & 80.8 & 70.7 & 75.6 & 78.6 & 73.0 & 78.8 & 81.4 \\
        \bf \shortname$^\dagger$ & \bf Ours & \bf{87.4} & \bf{88.4} & \bf{89.0} & \bf 76.4 & \bf 82.6 & \bf 84.9 & \bf 81.8 & \bf 83.3 & \bf 84.0 & \bf 79.2 & \bf 83.3 & \bf 84.5 \\
        \hline
        \end{tabular}
    \end{adjustbox}
    \caption{
    Unsupervised hashing results.
    *: Originally reported. 
    $\dagger$: Using contrastive learning.
    }
    \vspace{-0.3cm}
    \label{tab:retrieval_performance_unsupervised}
\end{table*}

\begin{figure}[t]
    \centering
    \begin{subfigure}[b]{0.49\linewidth}
        \centering
        \includegraphics[width=0.49\linewidth, keepaspectratio=True]{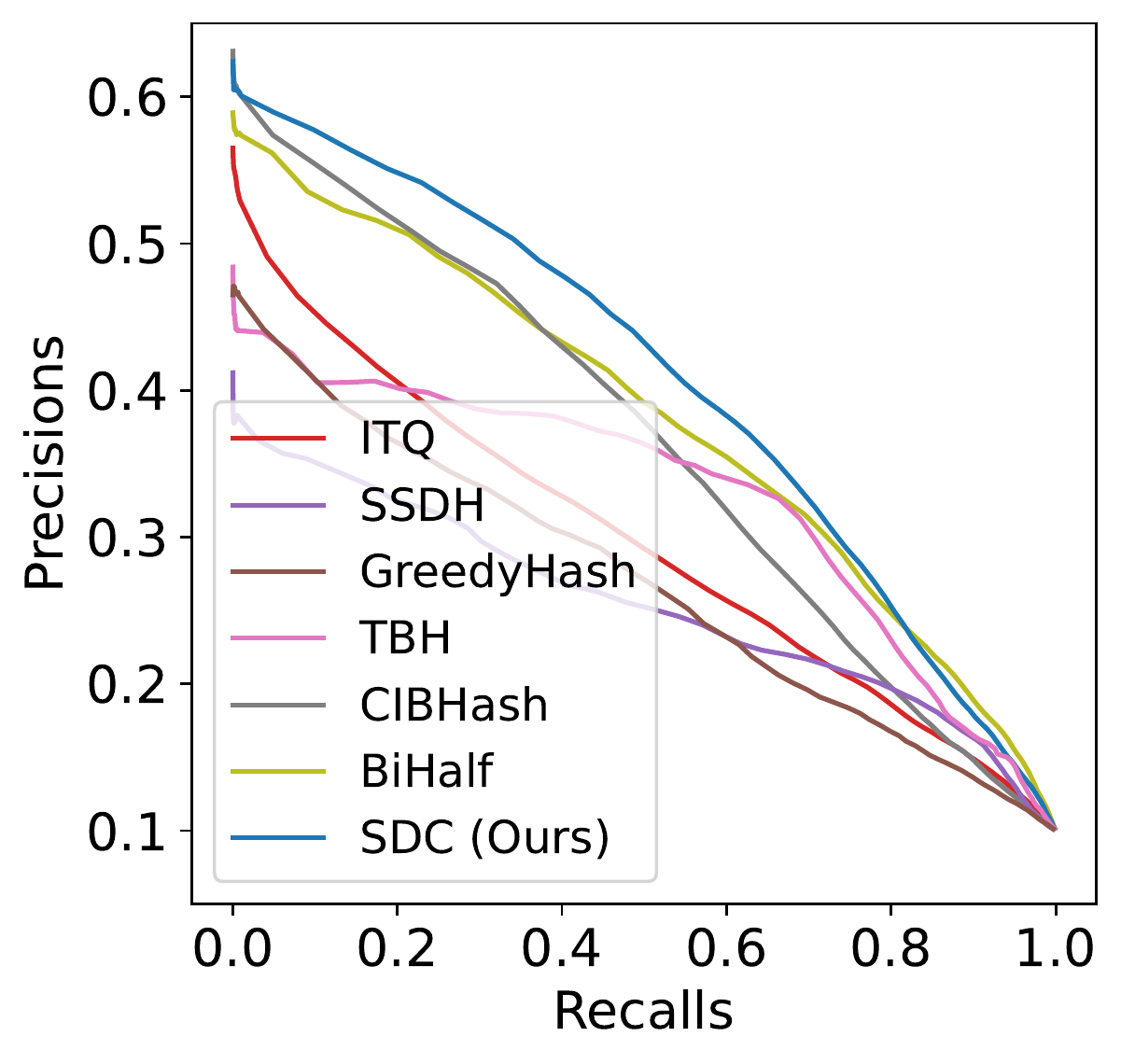}
        \includegraphics[width=0.49\linewidth, keepaspectratio=True]{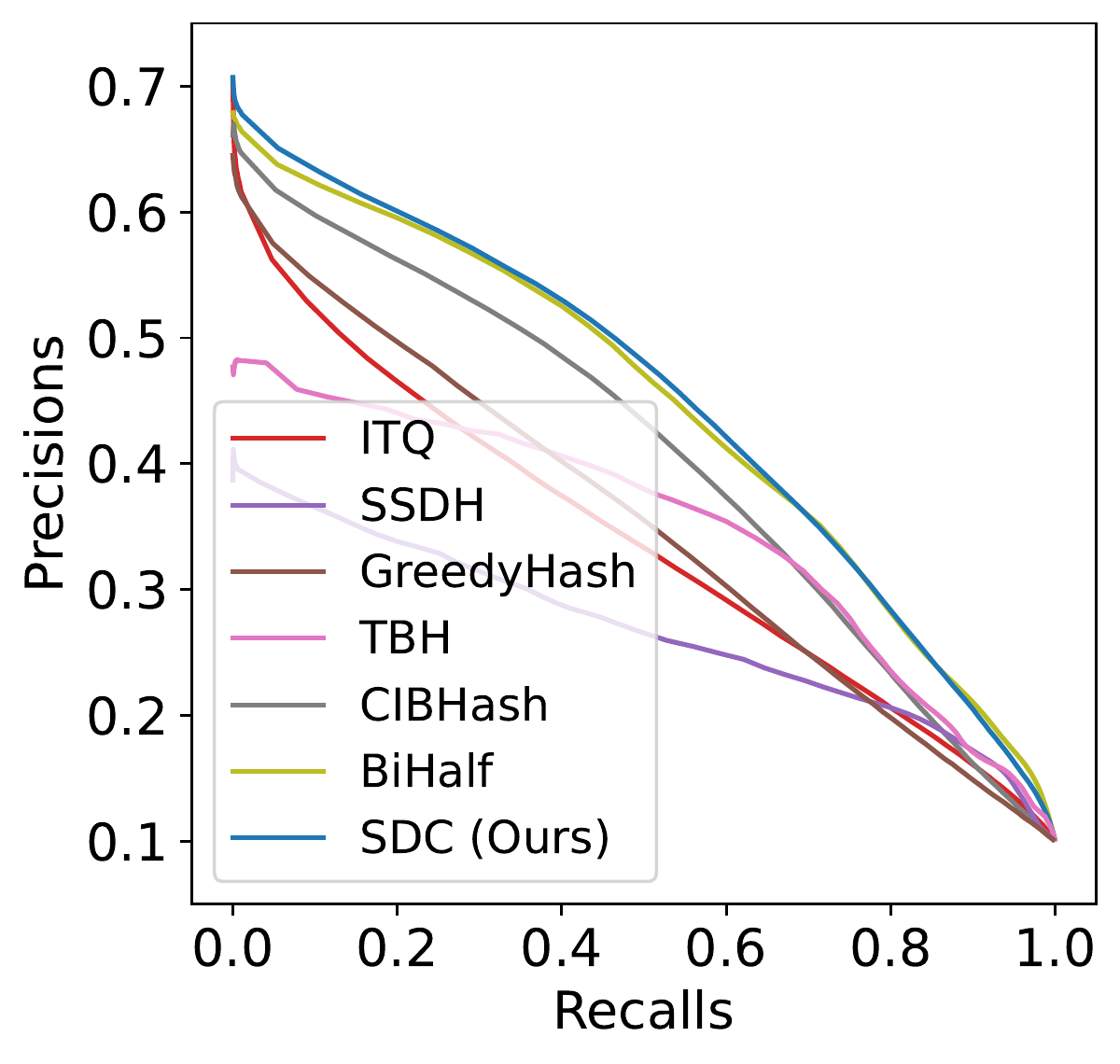}
        \caption{CIFAR10}
    \end{subfigure}
    \begin{subfigure}[b]{0.49\linewidth}
        \centering
        \includegraphics[width=0.49\linewidth, keepaspectratio=True]{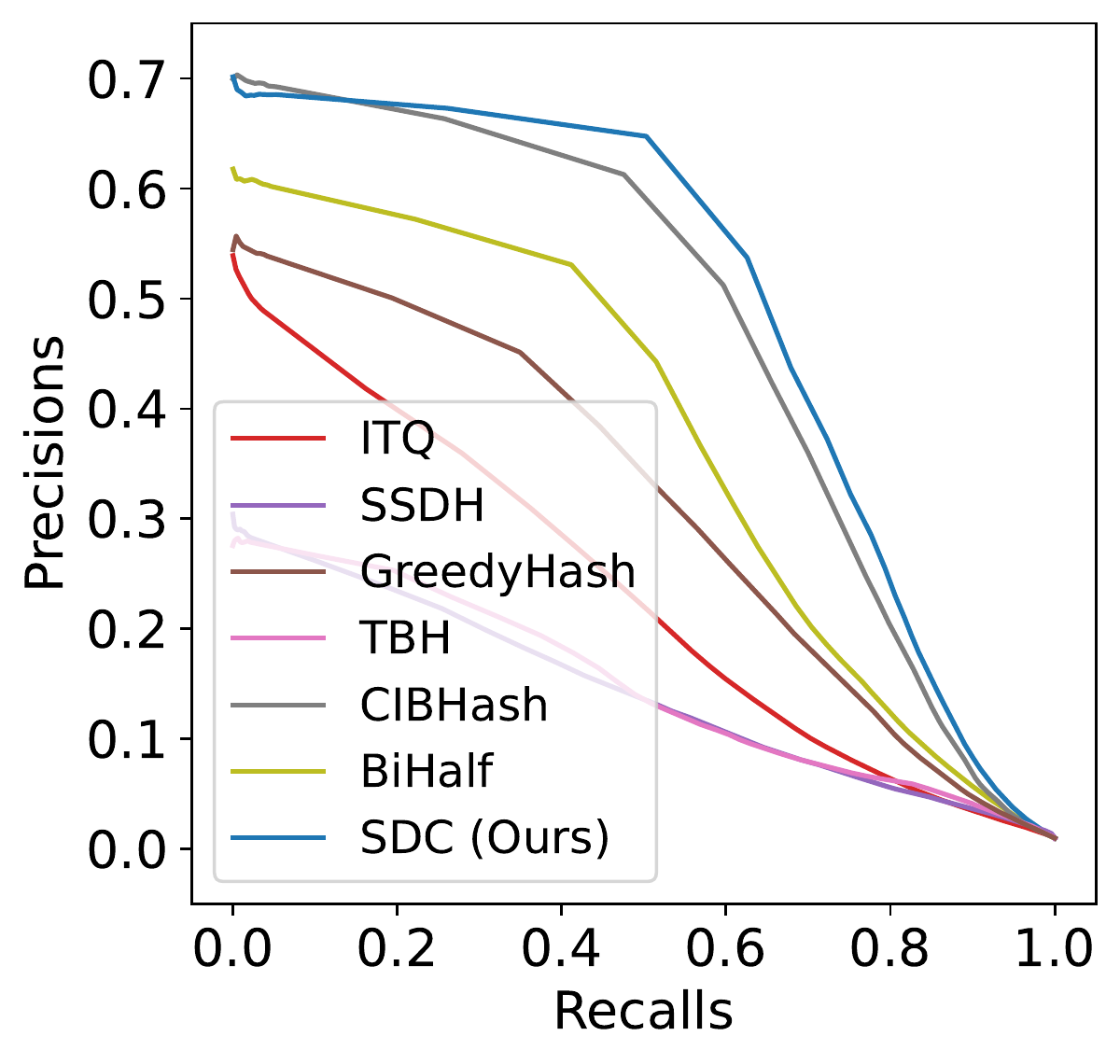}
        \includegraphics[width=0.49\linewidth, keepaspectratio=True]{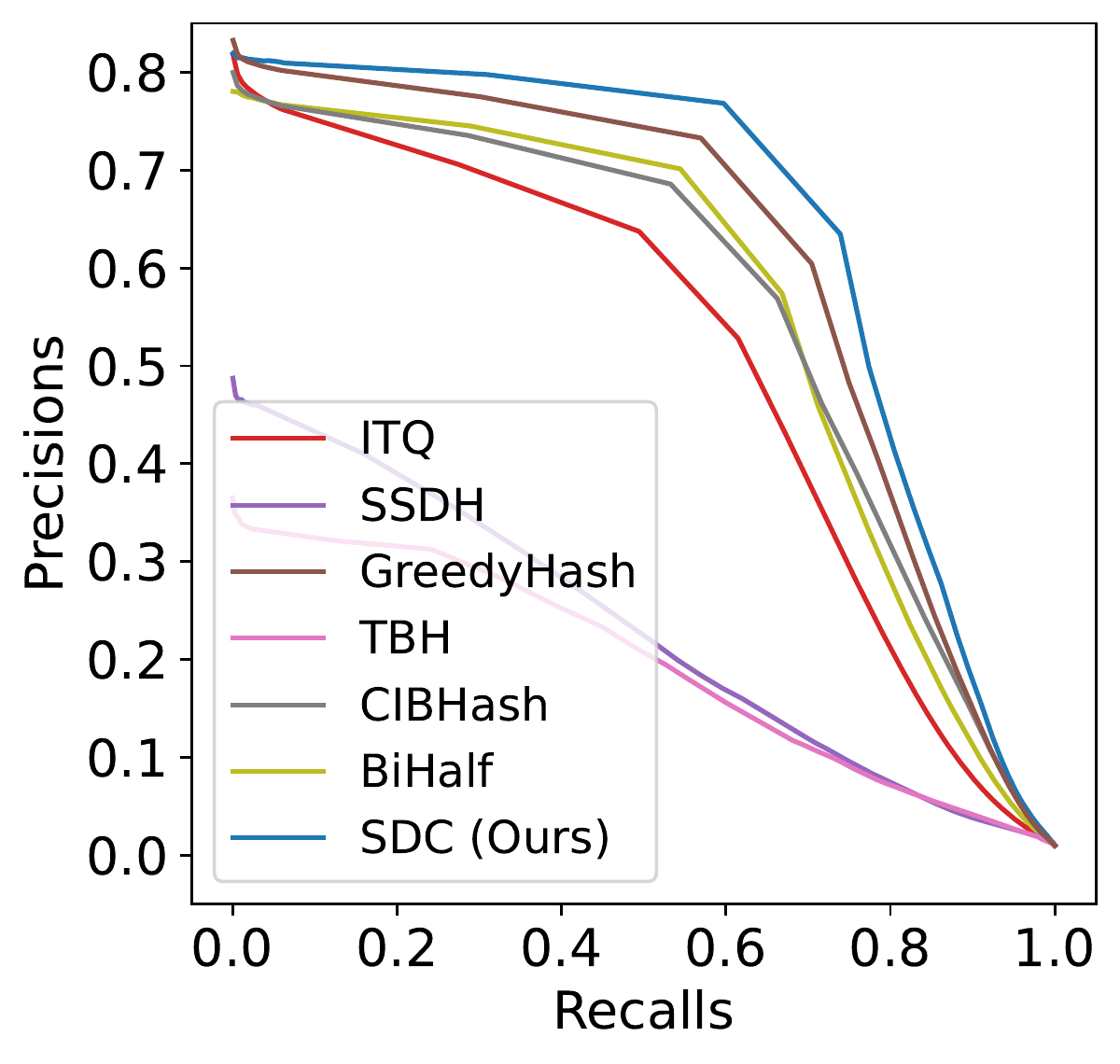}
        \caption{ImageNet100}
    \end{subfigure}
    \vspace{-0.3cm}
    \caption{PR curves on CIFAR10 and ImageNet100. For each dataset, the left and right correspond to 16-bits and 64-bits hash codes. Feature: VGG-16.}
    \label{fig:pr_curve}
\end{figure}


\vspace{0.1cm}
\noindent \textbf{Instance-level retrieval results.} 
We also evaluate our model on instance-level image retrieval tasks. 
%
\changed{For training efficiency,
we turn off the contrastive loss (\ie, $\lambda_{\text{cl}}=0$)}.
We follow the evaluation protocol of \cite{ortho2021jk}\footnote{Please see supplementary material for implementation details.}.
We use three datasets, namely GLDv2 \cite{gldv22020weyand}, $\mathcal{R}$Oxf, and $\mathcal{R}$Paris  \cite{roxfparis2018filip}.
Note, due to no training data with $\mathcal{R}$Oxf and $\mathcal{R}$Paris, we use the training set of
\begin{wraptable}{l}{0.5\textwidth}
    \centering
    \begin{adjustbox}{max width=\linewidth}
        \begin{tabular}{|l|cc|cc|cc|}
        \hline
        \multirow{2}{*}{Methods} & \multicolumn{2}{c|}{GLDv2} & \multicolumn{2}{c|}{$\mathcal{R}$Oxf} & \multicolumn{2}{c|}{$\mathcal{R}$Paris} \\
        \cline{2-7}
        & 128 & 512 & 128 & 512 & 128 & 512 \\
        \hline
        ITQ \cite{itq2012gong} & 5.2 & 11.3 & 1.6 & 5.4 & 4.8 & 12.3 \\
        GreedyHash \cite{greedyhash2018su} & 3.8 & 7.9 & 15.8 & 34.2 & 34.9 & 52.8 \\
        BiHalf \cite{bihalf2021li} & 4.0 & 6.7 & 20.2 & 33.3 & 42.0 & 52.0 \\
        \bf \shortname{} (Ours) & \textbf{6.3} & \textbf{12.1} & \textbf{27.1} & \textbf{40.8} & \textbf{50.3} & \textbf{63.8} \\
        \hline
        \em Original features
        & \multicolumn{2}{c|}{13.8} & \multicolumn{2}{c|}{51.0} & \multicolumn{2}{c|}{71.5} \\
        \hline
        \end{tabular}
    \end{adjustbox}
    \caption{Instance-level image retrieval results of representative unsupervised hashing methods on GLDv2, $\mathcal{R}$Oxf-\textbf{Hard} and $\mathcal{R}$Paris-\textbf{Hard}. 
    Original features: 2048D R50-DELG features \cite{delg2020cao} with the cosine similarity.
    }
    \label{tab:instance_performance}
    \vspace{-0.5cm}
\end{wraptable}
 GLDv2 for model training for all datasets.
As shown in Table \ref{tab:instance_performance}, our {\shortname{}} still outperforms consistently the state-of-the-art similarity preservation based methods (\ie, GreedyHash \cite{greedyhash2018su}, and BiHalf \cite{bihalf2021li}) by a large margin.
This indicates that the superiority of our \shortname{} generalizes from coarse category retrieval to fine-grained instance retrieval, even in the presence of a distributional shift between the training and test sets.


\vspace{-0.2cm}
\subsection{Further Analysis} \label{sec:ablstudy}

\begin{figure*}[t]
    \centering
    \begin{subfigure}[b]{0.28\textwidth}
        \centering
        \includegraphics[keepaspectratio=True, scale=0.25]{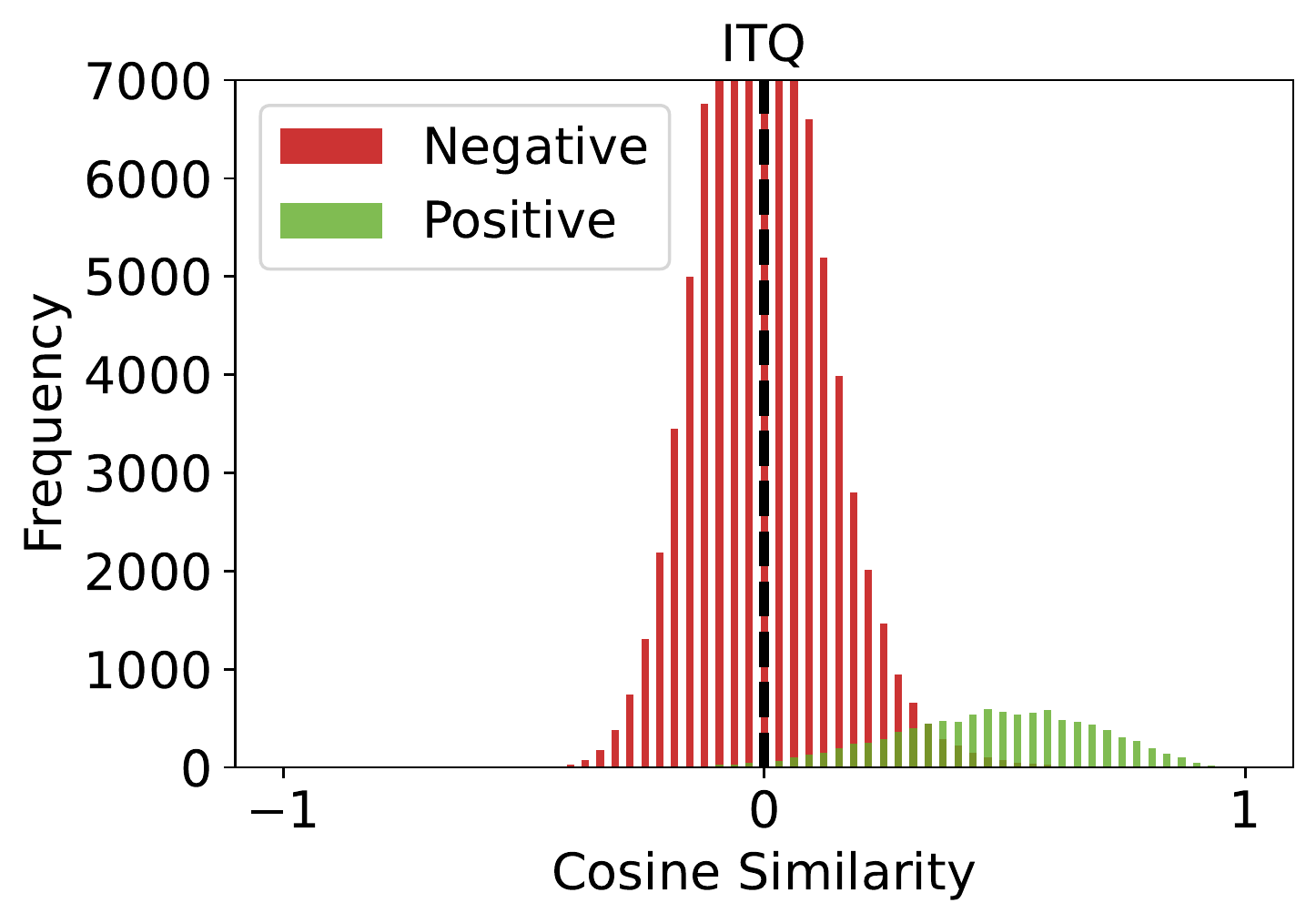}
        \caption{37.92\%}
        \label{fig:dist_histogram_itq}
    \end{subfigure}
    \hfill
    \begin{subfigure}[b]{0.23\textwidth}
        \centering
        \includegraphics[keepaspectratio=True, scale=0.25]{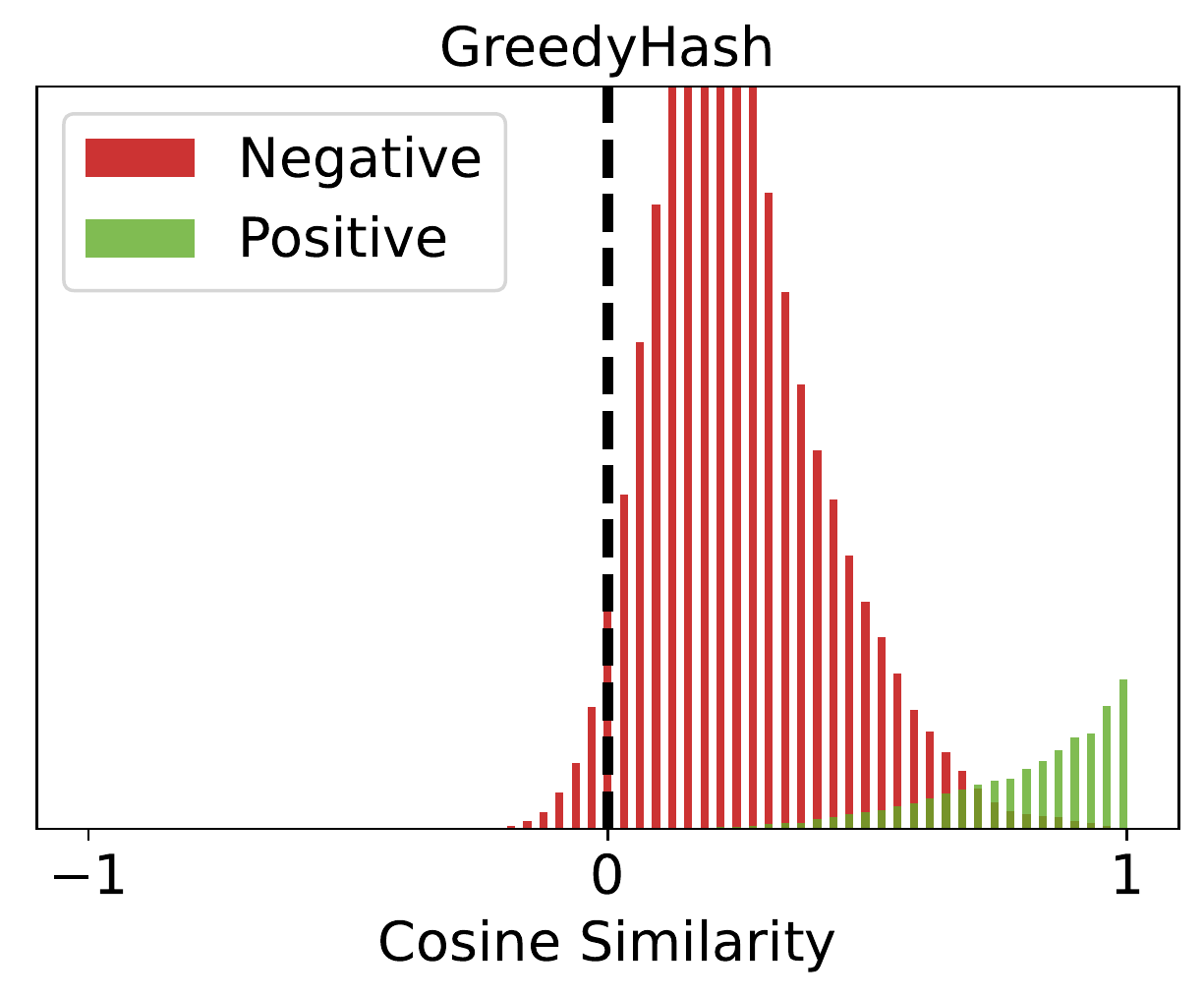}
        \caption{37.33\%}
        \label{fig:dist_histogram_bihalf}
    \end{subfigure}
    \hfill
    \begin{subfigure}[b]{0.23\textwidth}
        \centering
        \includegraphics[keepaspectratio=True, scale=0.25]{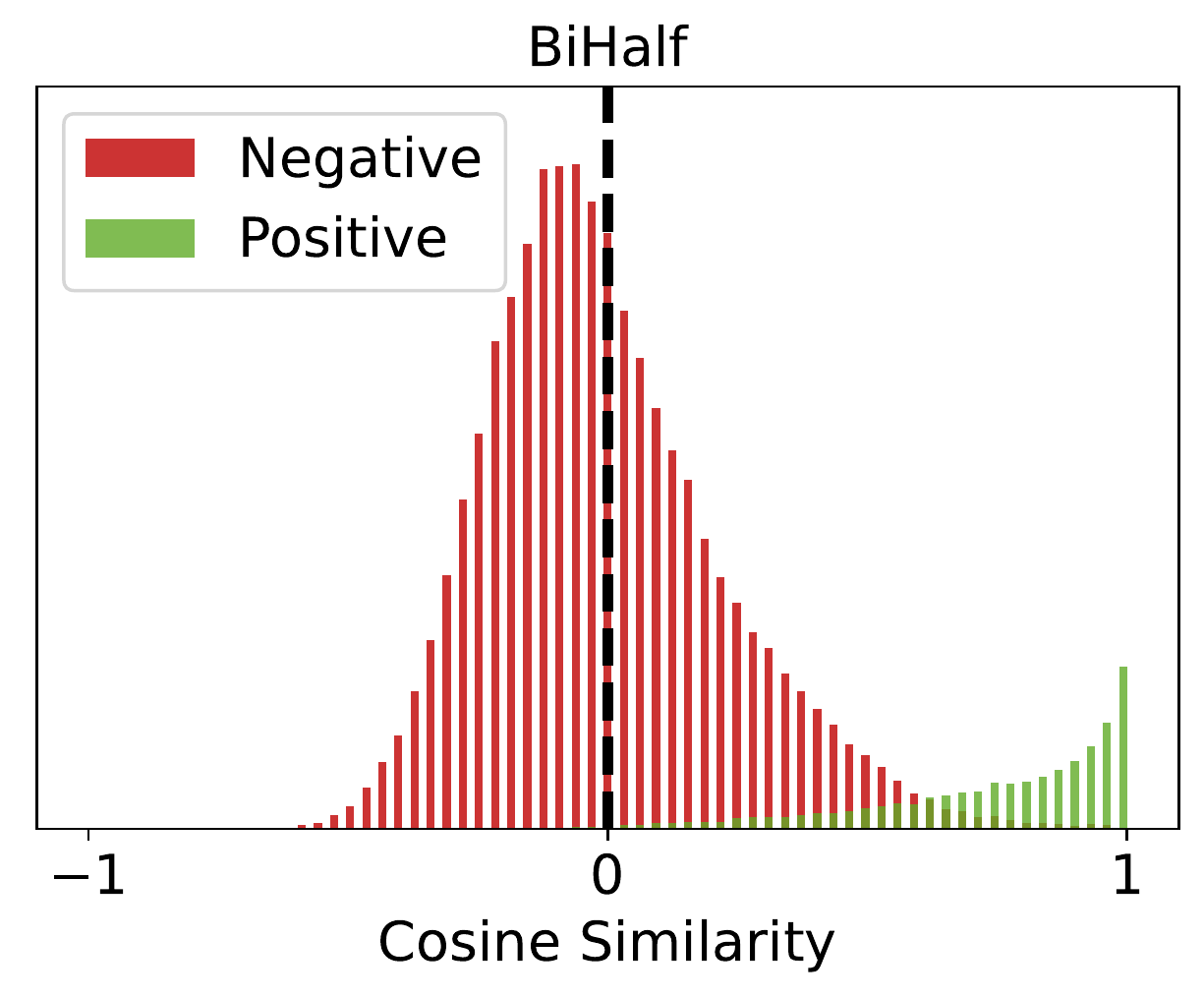}
        \caption{35.97\%}
        \label{fig:dist_collapse_are}
    \end{subfigure}
    \hfill
    \begin{subfigure}[b]{0.23\textwidth}
        \centering
        \includegraphics[keepaspectratio=True, scale=0.25]{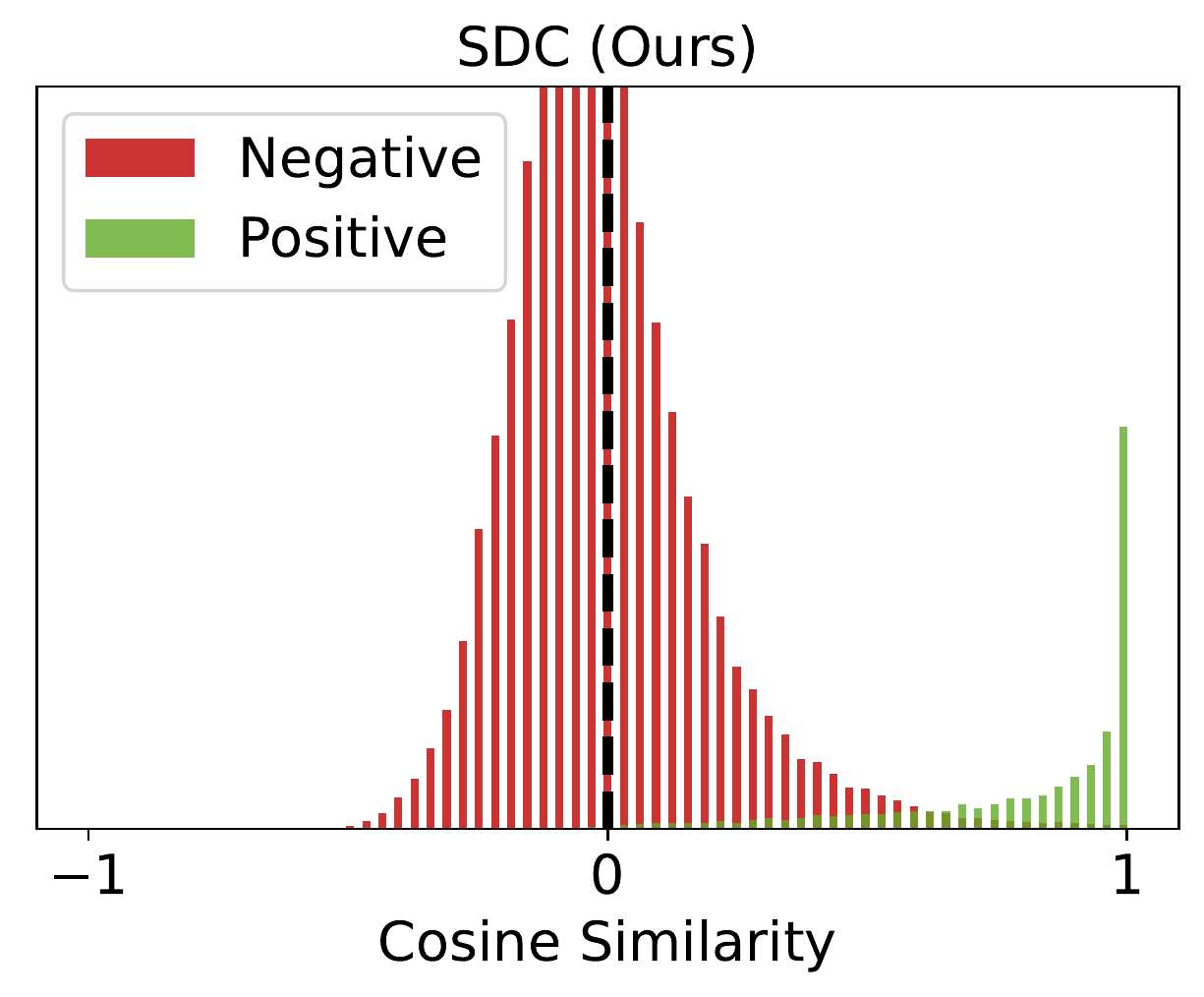}
        \caption{\bf 28.18\%}
        \label{fig:dist_collapse_arwe}
    \end{subfigure}
    \vspace{-0.2cm}
    \caption{
    Analysis of the similarity collapse problem on ImageNet100.
    We plot the hamming distance histograms for 10000 positive and 100000 negative random pairs with 64-bits hash codes. 
    For similarity collapse quantification, we use the intersection 
    between the two histograms as the metric, lower is better. Note that, the positive and negative labels are used for illustration purpose only, 
    but not used during model unsupervised training.
    }
    \label{fig:dist_histogram_all}
    \vspace{-0.3cm}
\end{figure*}

\begin{figure*}
    \centering
    \begin{subfigure}[b]{0.194\textwidth}
        \centering
        \includegraphics[scale=0.3, keepaspectratio=True]{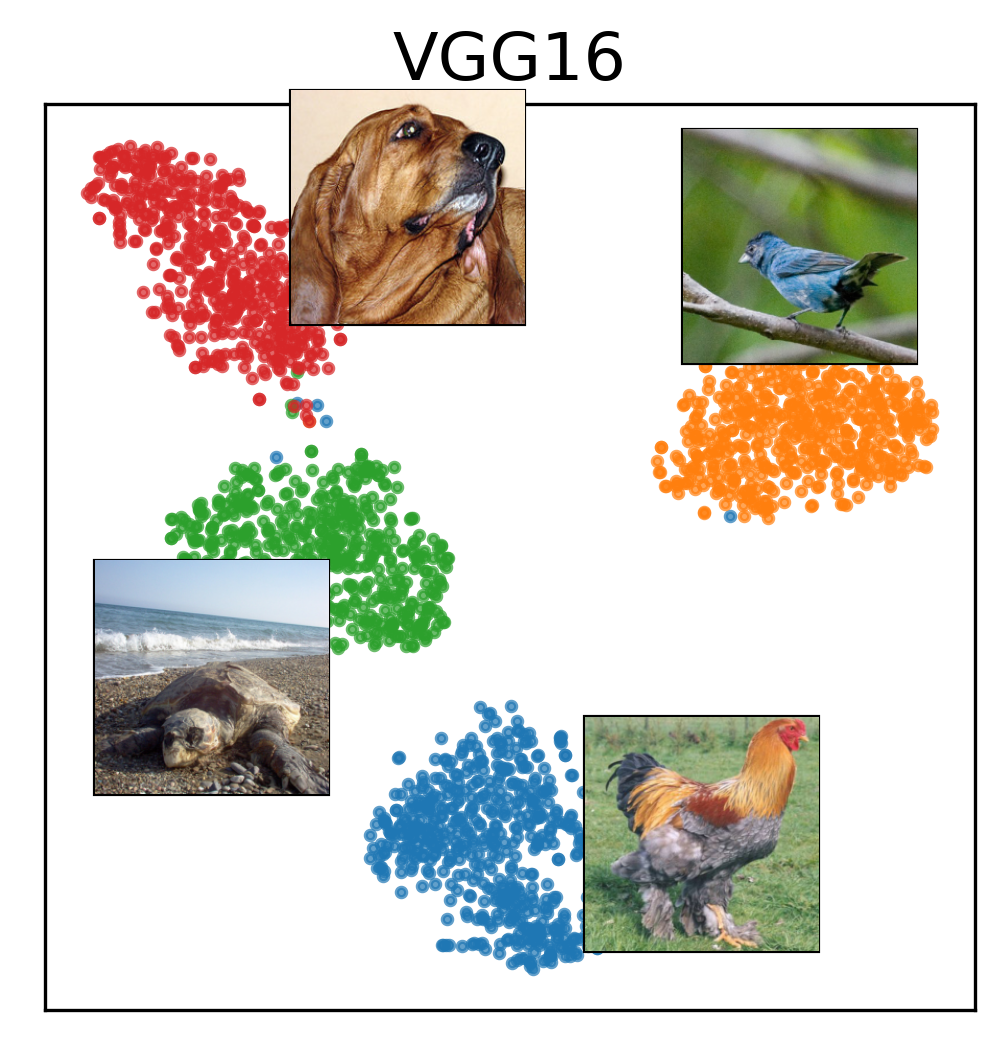}
        \caption{}
    \end{subfigure}
    \begin{subfigure}[b]{0.194\textwidth}
        \centering
        \includegraphics[scale=0.3, keepaspectratio=True]{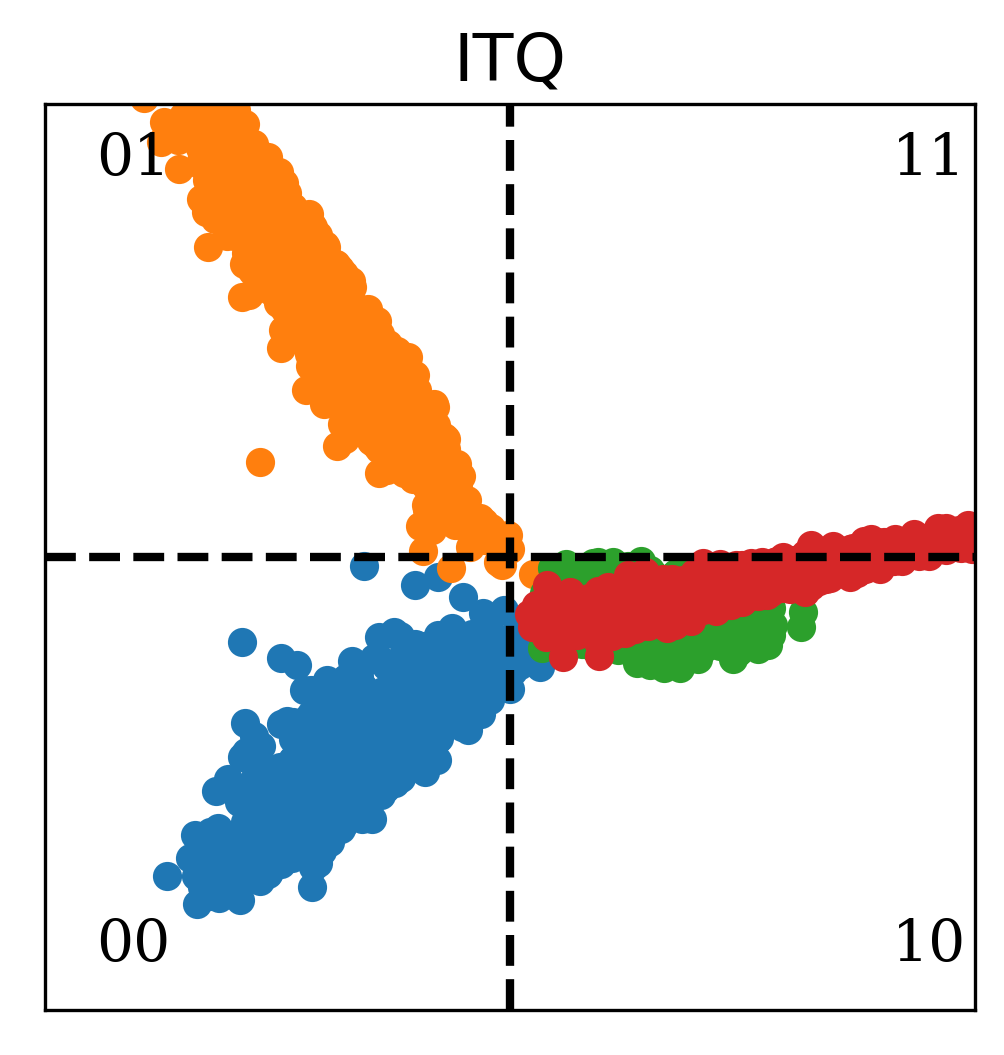}
        \caption{}
    \end{subfigure}
    \begin{subfigure}[b]{0.194\textwidth}
        \centering
        \includegraphics[scale=0.3, keepaspectratio=True]{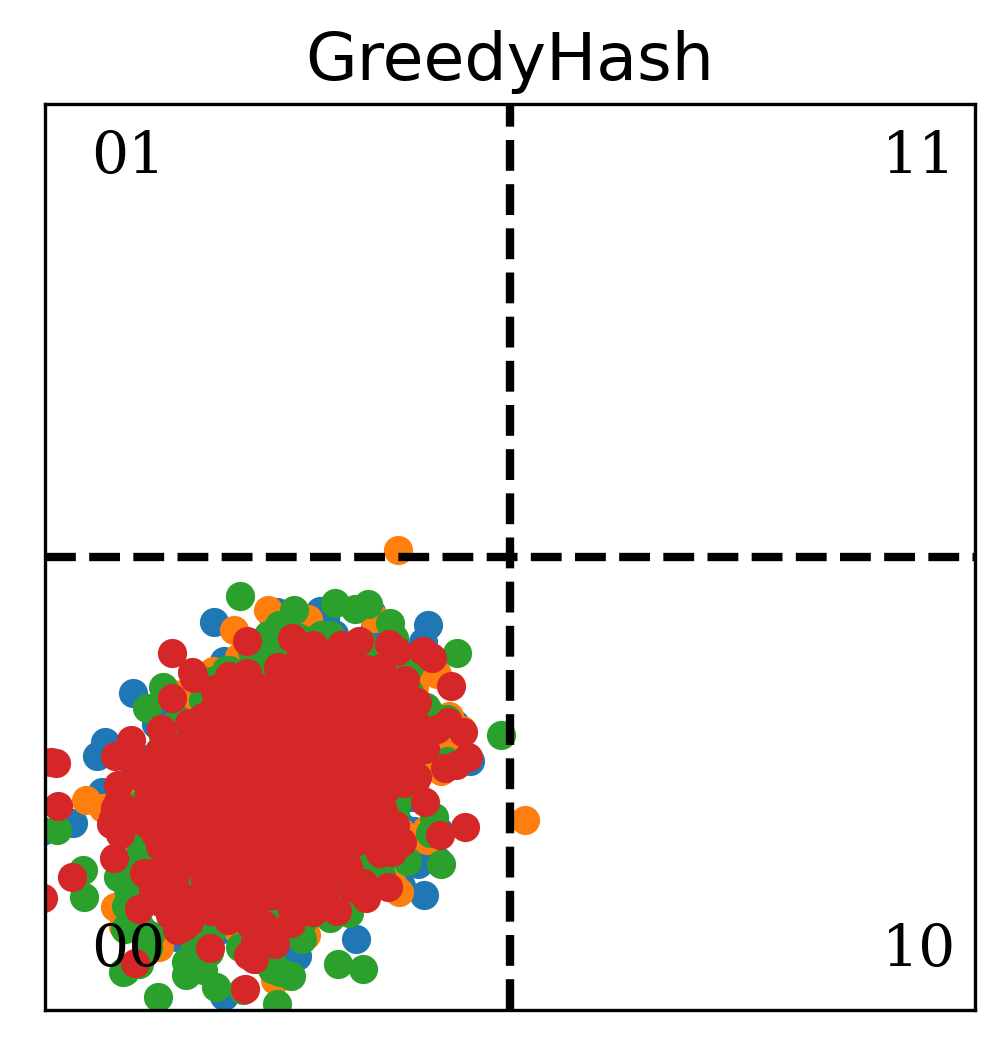}
        \caption{}
    \end{subfigure}
    \begin{subfigure}[b]{0.194\textwidth}
        \centering
        \includegraphics[scale=0.3, keepaspectratio=True]{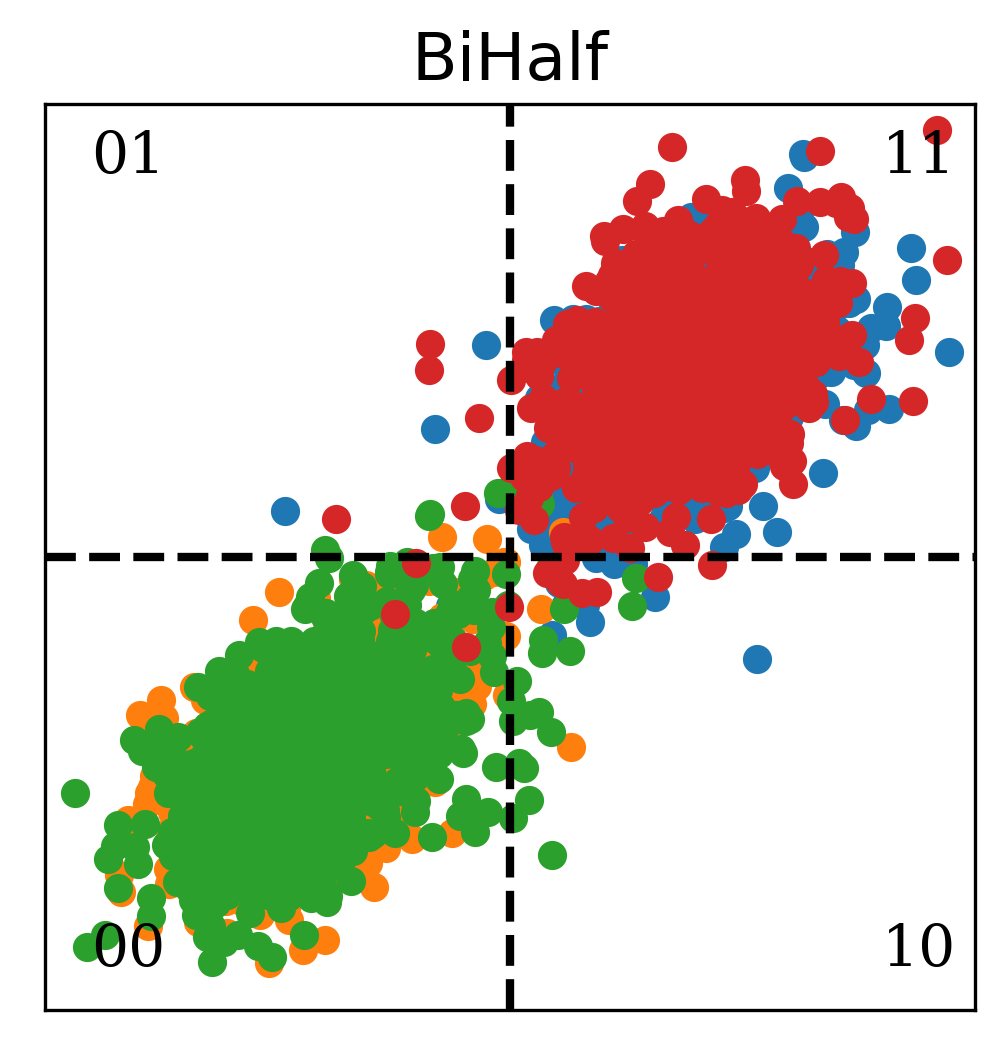}
        \caption{}
    \end{subfigure}
    \begin{subfigure}[b]{0.194\textwidth}
        \centering
        \includegraphics[scale=0.3, keepaspectratio=True]{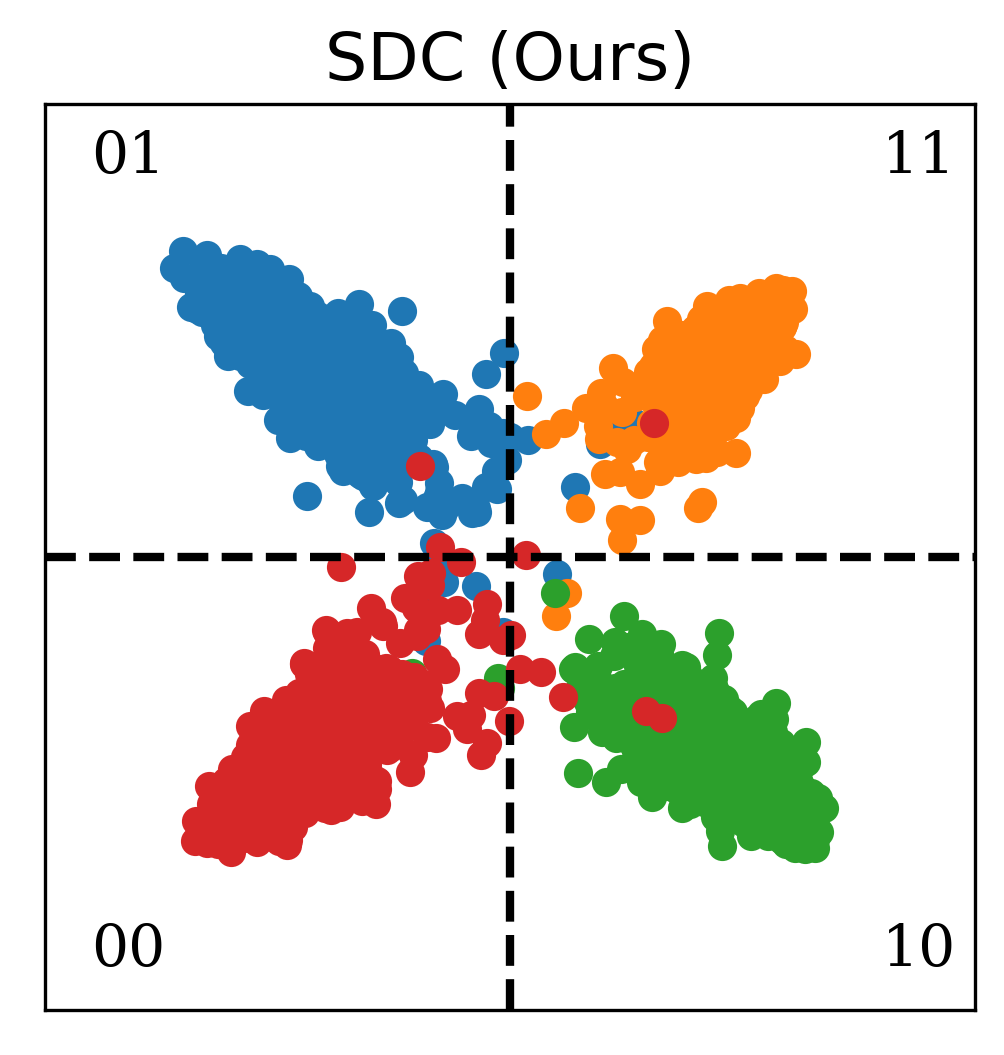}
        \caption{}
    \end{subfigure}
    \vspace{-0.6cm}
    \caption{\textbf{(a)} The t-SNE visualization of VGG16 features of 4 object classes of ImageNet100. \textbf{(b-e)} Continuous 2-bit codes before quantization derived by different unsupervised hashing methods. 
    Dotted lines denote the separation of Hamming space.}
    \label{fig:2bits}
    \vspace{-0.4cm}
\end{figure*}

\begin{figure*}[t]
    \centering
    \begin{subfigure}[c]{0.22\textwidth}
        \centering
        Query \\
        \includegraphics[scale=1, keepaspectratio=True]{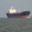}
    \end{subfigure}
    \hfill
    \begin{subfigure}[c]{0.38\textwidth}
        \centering
        \includegraphics[scale=0.3, keepaspectratio=True]{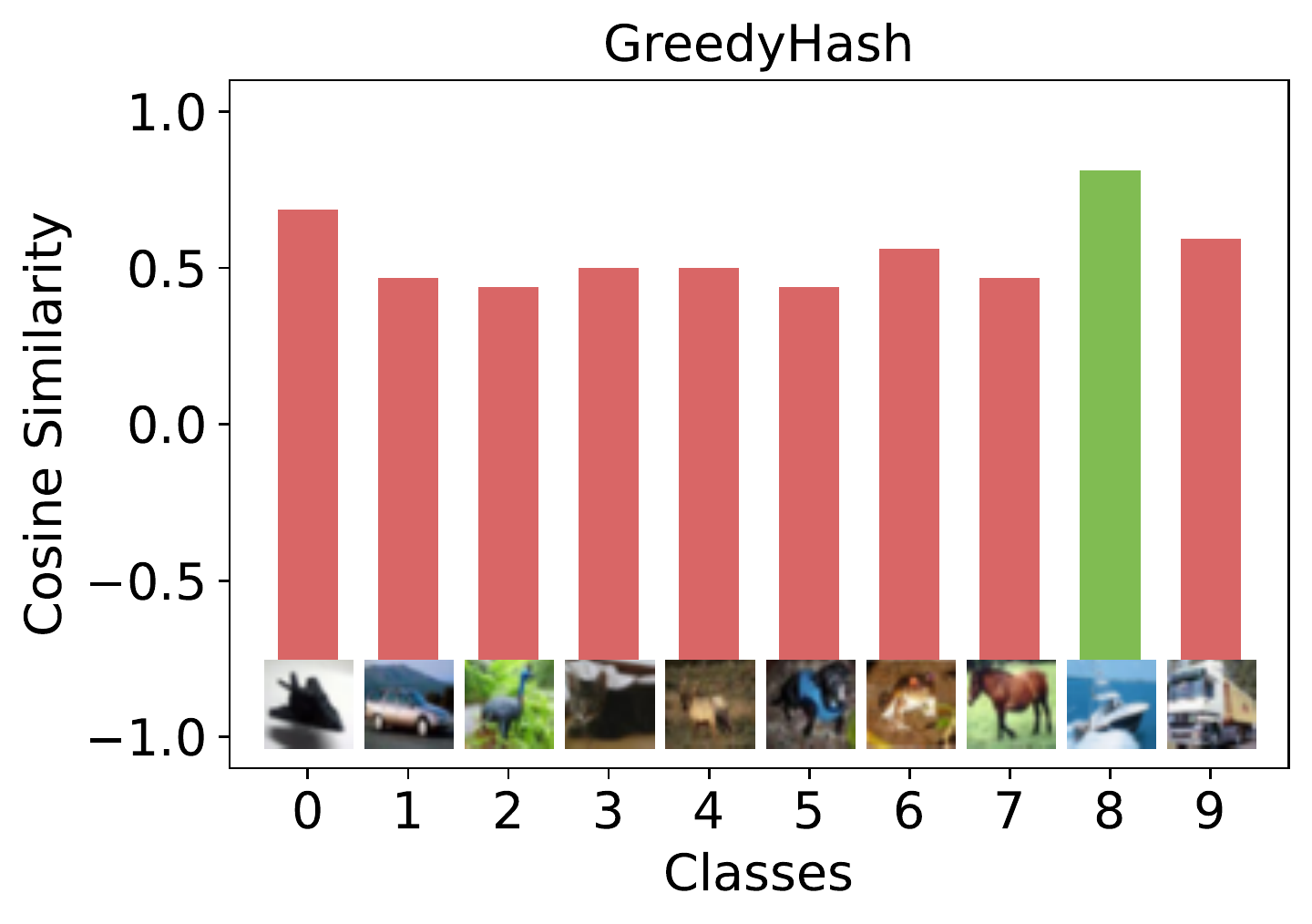}
    \end{subfigure}
    \begin{subfigure}[c]{0.38\textwidth}
        \centering
        \includegraphics[scale=0.3, keepaspectratio=True]{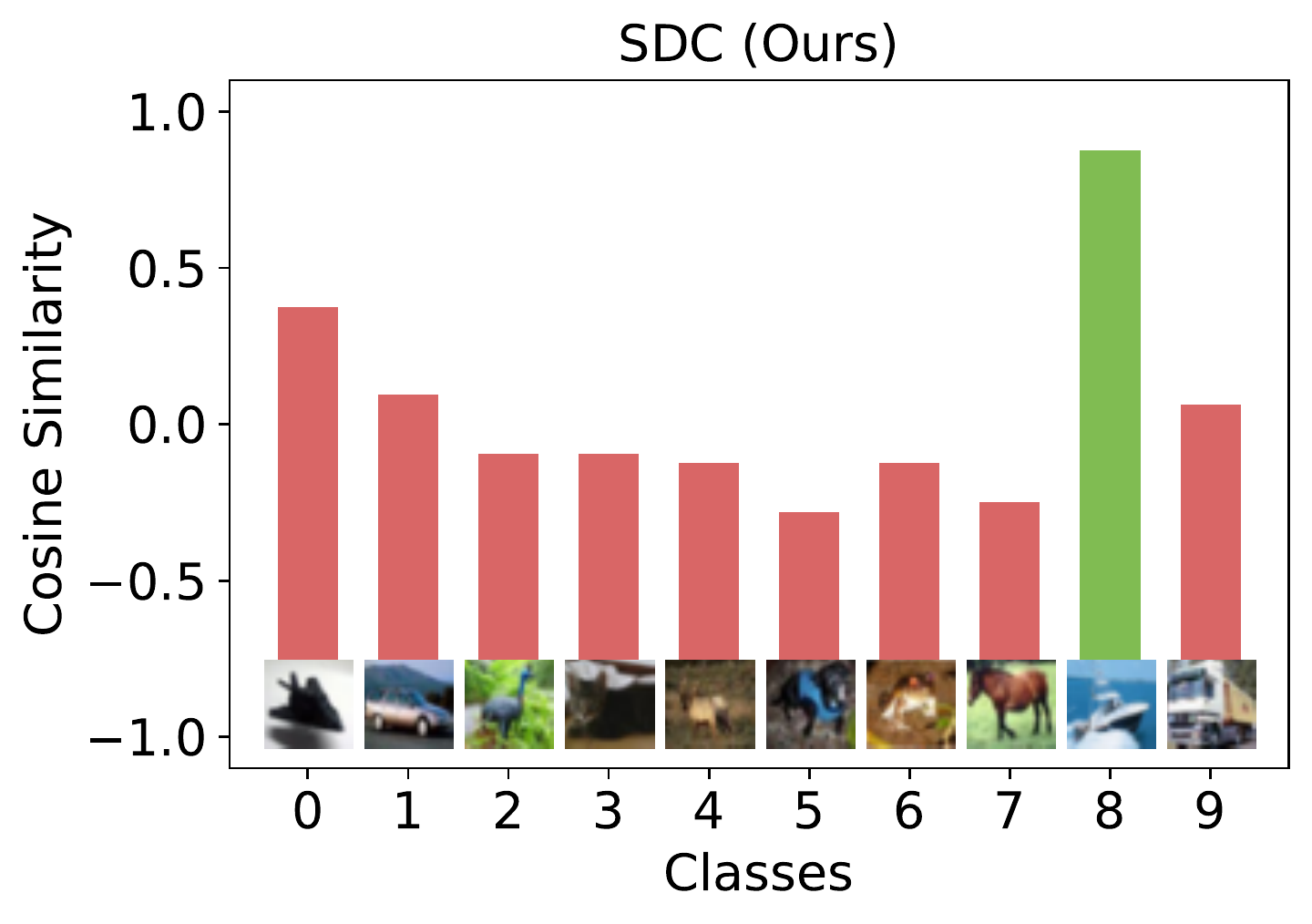}
    \end{subfigure}
    \begin{subfigure}[c]{0.22\textwidth}
        \centering
        Query \\
        \includegraphics[scale=1, keepaspectratio=True]{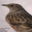}
    \end{subfigure}
    \hfill
    \begin{subfigure}[c]{0.38\textwidth}
        \centering
        \includegraphics[scale=0.3, keepaspectratio=True]{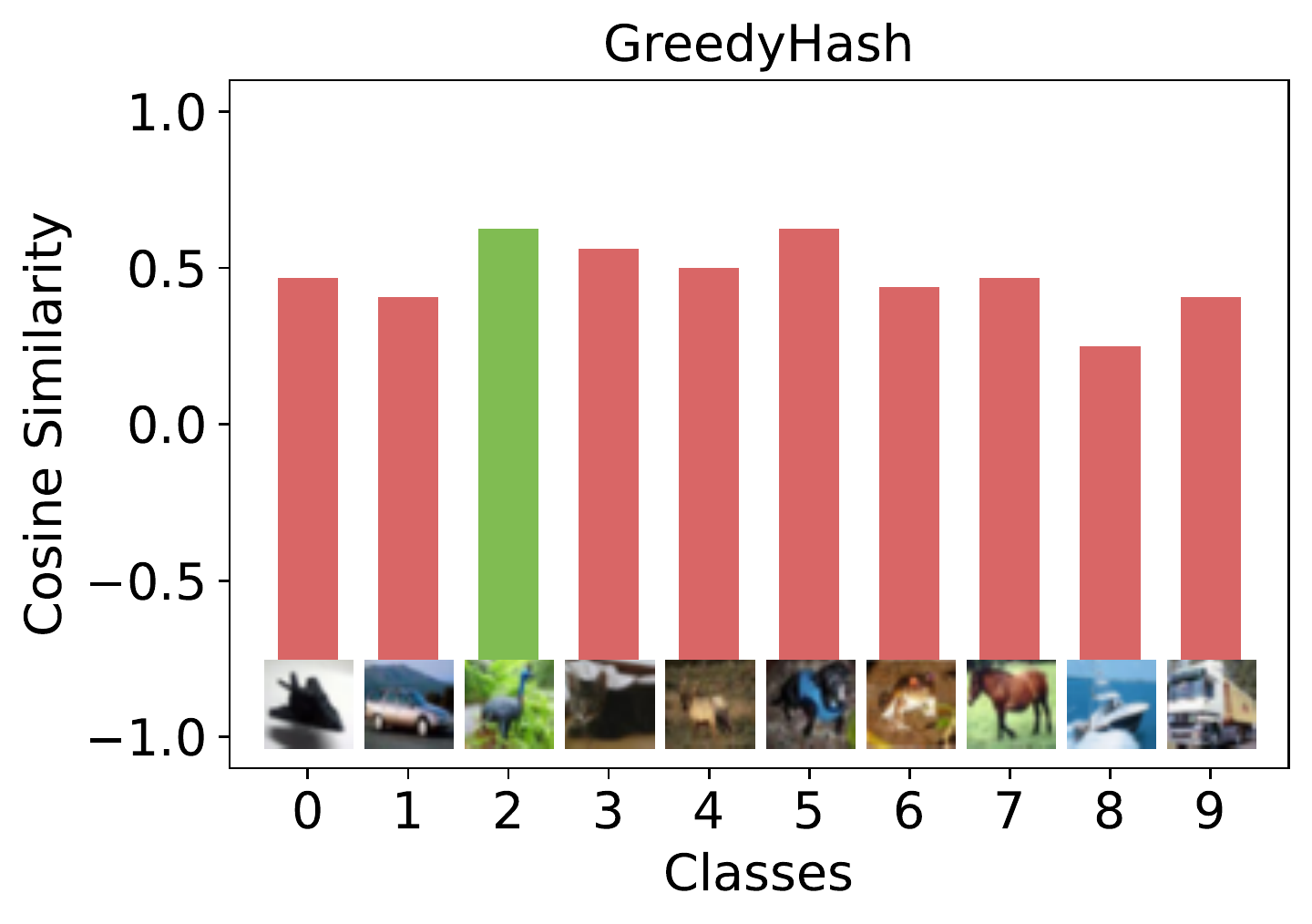}
    \end{subfigure}
    \begin{subfigure}[c]{0.38\textwidth}
        \centering
        \includegraphics[scale=0.3, keepaspectratio=True]{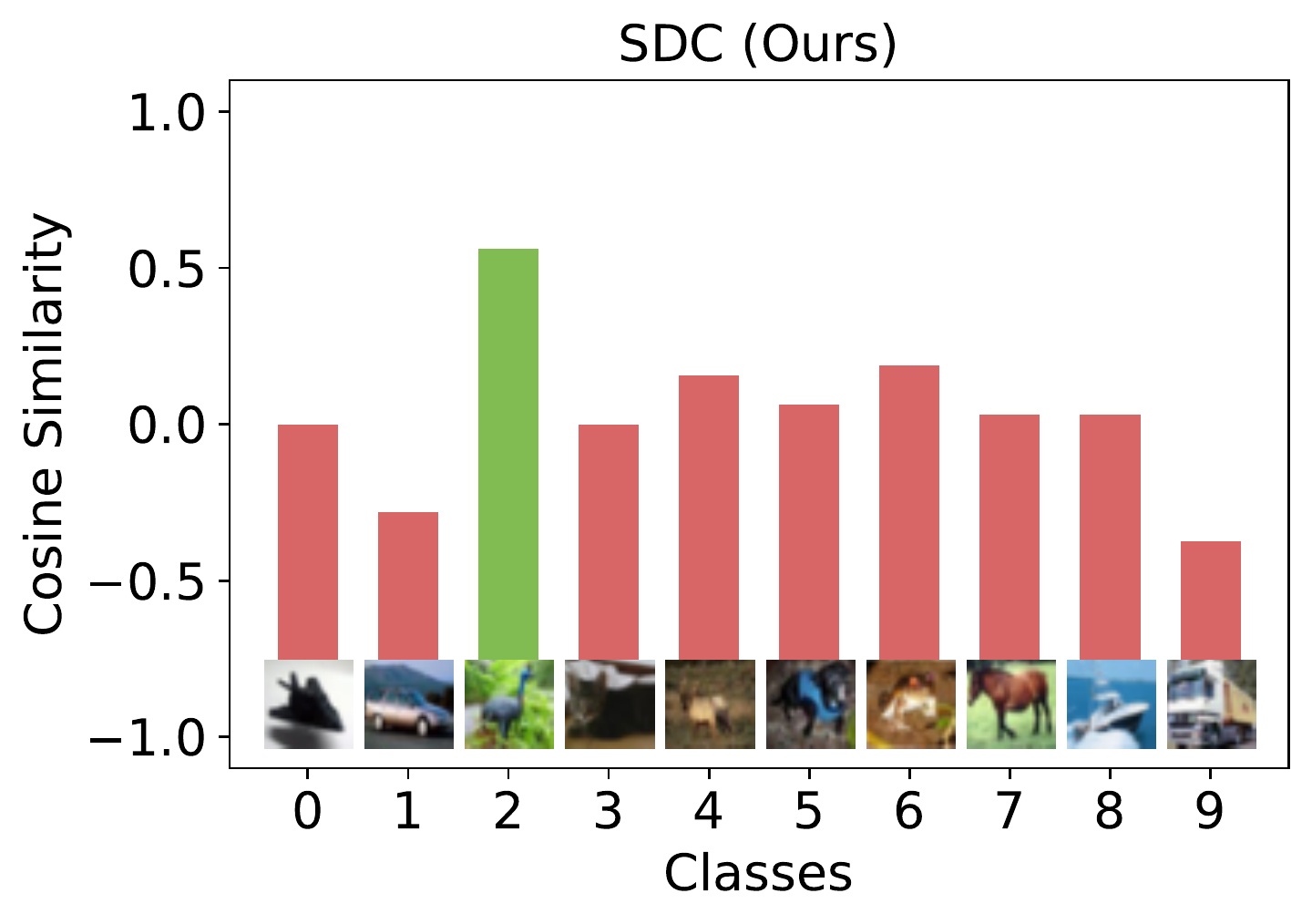}
    \end{subfigure}
    \caption{
    Two qualitative object image retrieval examples on CIFAR10.
    Green: Positive class;
    Red: Negative class.
     }
    \label{fig:queryexample}
    \vspace{-0.3cm}
\end{figure*}

\noindent \textbf{Similarity collapse analysis.} 
We first examine the similarity collapse problem.
We study three representative \changed{similarity preservation-based} hashing methods ({ITQ}, {GreedyHash}, {Bihalf}) in comparison with our {\shortname{}}.
To quantify this collapse, we compute the intersection
between the cosine similarity histogram of positive and negative pairs.
Higher intersection rates suggest worse collapses with lower discriminating ability.
We use 64-bits hash codes.
To compute the two histograms, we sample 10k positive and 100k negative random pairs of ImageNet100.
Fig.~\ref{fig:dist_histogram_all} presents the degree of  similarity collapse in the order of 
{ITQ} $>$ {GreedyHash} $>$ {BiHalf} $>$ {\shortname{}}.
This verifies again that our method is effective in alleviating this collapse problem.


\vspace{0.1cm}
\noindent \textbf{Hash code visualization.} 
We visualize continuous hash codes in a proof-of-concept setting.
Specifically, we examine the behaviour of ITQ, GreedyHash, BiHalf and \shortname{} in learning a 2-bits hash function 
over 4 object classes (\texttt{cock}, \texttt{indigo-bunting}, \texttt{loggerhead}, \texttt{bloodhound}) from ImageNet100.
We use the VGG-16 features.
We observe from Fig. \ref{fig:2bits} that whilst the original features are already well separable, different methods behave differently.
For example, simply preserving the original similarity, GreedyHash collapses the similarity scores completely across all the classes. With a code balance layer on top, BiHalf partly reduces the degree of collapsing to two groups. Through aligning the similarity distribution with a calibration distribution, our SDC solves this collapsing problem well, even further separating the originally confusing two classes ({\texttt{cock} and \texttt{bloodhound}}).
%
This validates the unexceptional potential of SDC in improving the retrieval performance over previous methods.

\vspace{0.1cm}
\noindent \textbf{Qualitative evaluation.} 
For visual analysis, we provide a couple of image retrieval examples
on CIFAR10. 
It is evident in Fig.~\ref{fig:queryexample} that 
our SDC can identify the positive class more confidently 
with a more distinctive separation between positive and negative classes compared to GreedyHash.
This indicates the superior discrimination ability of our similarity distribution calibration idea in unsupervised hashing.





\section{Conclusion} \label{sec:conclusion}

We have presented a simple yet effective {\em \modelname} (\shortname{}) method for unsupervised hashing.
This is particularly designed to alleviate the largely ignored \textit{similarity collapse} problem suffered by the existing similarity preservation-based unsupervised hashing methods.
Concretely, we minimize the Wasserstein distance between the distribution of Hamming similarities and a calibration distribution with a sufficient capacity range.
As a result, the low similarity range of hash code can be better exploited for improved discriminating ability.
Extensive experiments on both coarse and fine-grained image retrieval tasks validated the advantage of our method over the state-of-the-art alternatives.

\bibliography{main}
\end{document}